\documentclass[sigconf, nonacm]{acmart}
\usepackage{mathtools}
\usepackage{algorithm}
\usepackage[noend]{algpseudocode}
\usepackage{subfigure}
\usepackage{makecell}
\usepackage{multirow}
\usepackage{amsmath}
\usepackage{amsfonts}
\usepackage{dsfont}
\usepackage{color}
\newtheorem{definition}{\bf Definition}

\newtheorem{example}{\bf Example}
\newtheorem{property}{\bf Property}
\newtheorem{strategy}{\bf Strategy}
\newtheorem*{theorem*}{Research Problem}

\DeclarePairedDelimiter\ceil{\lceil}{\rceil}

\newcommand{\Xiao}[1]{(\textcolor{purple}{Xiao: {#1}})}
\newcommand\vldbdoi{XX.XX/XXX.XX}

\newcommand\vldbavailabilityurl{URL_TO_YOUR_ARTIFACTS}
\newcommand\vldbpagestyle{plain} 

\begin{document}
\title{Skeleton-Guided Learning for Shortest Path Search}

\author{Tiantian Liu$^{\dagger}$ Xiao Li$^{\S}$ Huan Li$^{\ddagger}$ Hua Lu$^{\dagger}$ Christian S. Jensen$^{\dagger}$ Jianliang Xu$^{\natural}$}
\affiliation{%
\institution{
{\fontsize{10}{10}\selectfont $^{\dagger}$}Department of Computer Science, Aalborg University, Denmark\\
{\fontsize{10}{10}\selectfont $^{\S}$}Department of Computer Science, IT University of Copenhagen, Denmark \\
{\fontsize{10}{10}\selectfont $^{\ddagger}$}College of Computer Science and Technology, Zhejiang University, China \\
{\fontsize{10}{10}\selectfont $^\natural$}Department of Computer Science, Hong Kong Baptist University, Hong Kong\\
{\fontsize{7}{7}\selectfont\ttfamily\upshape} }$^{\dagger}$\{liutt, luhua, csj\}@cs.aau.dk, 
$^{\S}$xliq@itu.dk,
$^{\ddagger}$lihuan.cs@zju.edu.cn,  $^\natural$xujl@comp.hkbu.edu.hk}







\begin{abstract}
Shortest path search is a core operation in graph-based applications, yet existing methods face important limitations. Classical algorithms such as Dijkstra’s and A* become inefficient as graphs grow more complex, while index-based techniques often require substantial preprocessing and storage. Recent learning-based approaches typically focus on spatial graphs and rely on context-specific features like geographic coordinates, limiting their general applicability. 
We propose a versatile learning-based framework for shortest path search on generic graphs, without requiring domain-specific features. At the core of our approach is the construction of a skeleton graph that captures multi-level distance and hop information in a compact form. A Skeleton Graph Neural Network (SGNN) operates on this structure to learn node embeddings and predict distances and hop lengths between node pairs. These predictions support $\textsc{LSearch}$, a guided search algorithm that uses model-driven pruning to reduce the search space while preserving accuracy.
To handle larger graphs, we introduce a hierarchical training strategy that partitions the graph into subgraphs with individually trained SGNNs. This structure enables $\textsc{HLSearch}$, an extension of our method for efficient path search across graph partitions.
Experiments on five diverse real-world graphs demonstrate that our framework achieves strong performance across graph types, offering a flexible and effective solution for learning-based shortest path search.
\end{abstract}

\maketitle

\pagestyle{\vldbpagestyle}


\section{Introduction}
\label{sec:intro}

 With the increasing digitalization of societal processes, numerous complex real-world graphs are being collected and analyzed. Graphs arise across a variety of domains, including social networks, web networks, power grids, road infrastructures, and biological systems.
 
 Finding the shortest path in a graph is classical functionality and has numerous applications.
 For example, in social networks, shortest path search is commonly used to determine the proximity between two individuals. Shortest path distances can be used to quantify the influence of a person, or they can help assess community membership or the evolution of a community~\cite{zhao2011efficient,tian2012shortest,newman2001scientific}.
 In a web network, the shortest path between two pages is important for finding a page with higher relevance to the page that is currently being served to a user~\cite{khater2014improving,yin2014shortest}.
 In power grids, shortest paths are crucial when performing reconfiguration of power system components~\cite{jelenius2004graph,lin2018shortest}.
 In road networks, shortest paths are used by drivers or pedestrians when they plan an appropriate path in real life~\cite{walter2006shortest,bai2016pedestrian}.
 In biological networks, shortest paths can help analyze connectivity or interaction between two components~\cite{assenov2008computing,pavlopoulos2011using}.
 In addition, shortest path functionality is foundational and can be leveraged to support more complex path search  functionality. 
 Therefore, efficient support for finding shortest paths offers substantial benefits across numerous domains.
 
Due to the significance of the shortest path problem, it has attracted substantial attention, and many solutions are available. Classical solutions such as Dijkstra algorithm~\cite{dijkstra1959note}, the Bellman-Ford algorithm~\cite{bellman1958routing,ford1956network}, the A* search algorithm~\cite{hart1968formal}, and the Floyd-Warshall algorithm~\cite{floyd1962algorithm} are inefficient for large and complex graphs. 
Traditional techniques for speeding up these methods, e.g., index-based methods~\cite{sanders2005highway, geisberger2008contraction, hassan2016graph,wang2016effective,ouyang2018hierarchy, ouyang2020efficient, qiu2022efficient}, landmark-based methods~\cite{francis2001idmaps,ng2002predicting,kleinberg2004triangulation,kriegel2008hierarchical,potamias2009fast,gubichev2010fast,qiao2011querying,qiao2012approximate, akiba2013fast}, bidirectional methods~\cite{luby1989bidirectional,geisberger2008contraction,vaira2011parallel,cabrera2020exact}, and graph compression~\cite{qiao2011querying}, either incur considerable pre-processing time or require substantial  index storage space. 
Motivated by its success in many settings, deep learning has also been adopted for path search~\cite{qi2020learning, yin2021learning, huang2021learning, zhao2022rne}. However, existing solutions generally target one specific scenario. In particular, graphs come in the form of spatial graphs or in the form of complex graphs that model social networks, web networks, power grids, or biological networks ~\cite{akiba2013fast, zhang2022shortest}. Most of the aforementioned methods are designed solely for road networks which require context information as features.

In this paper, we aim to enable learning-based shortest path search applicable to all types of graphs.
This is non-trivial due to several difficulties.
First, graphs from different fields exhibit different structural characteristics. For example, some graphs consist of star-topology subgraphs, and some graphs have a web-topology with multiple layers of rings interconnected. Second, it is challenging to capture complex relationships within a graph and to predict shortest distances. Third, the growth of graph data has recently been explosive, making it difficult to scale shortest path search to such large graphs.
 
We propose a framework that encompasses several techniques. 
First, it includes a method for efficiently capturing the distance and hop information in a generic graph. 
In this method, each vertex is assigned a specific set of labels.
These labels are not arbitrary; they are carefully determined using a base value and a series of scalable factors, making it both flexible and concise for representing the hop distances of vertices (to be detailed in~Section~\ref{sec:skeleton_graph}).
Such labels are then connected to form the ``skeleton graph'' of the original graph. 
This skeleton graph presents a more coherent and structured overview of the graph's architecture, simplifying the analysis and comprehension of the connections between nodes based on their distance and hop information.

Next, we provide a Skeleton Graph Neural Network (SGNN) for distance and hop length prediction for two original graph vertices. Unlike in Graph Neural Networks (GNN), which considers the original graph, we aggregate information using the message passing on a skeleton graph in SGNN. The aggregated information enables us to generate low-dimensional embeddings of vertices. This embedding method does not rely on context information such as longitude and latitude, or road categories in road networks, making SGNNs broadly applicable to many kinds of graphs. Using the embeddings and SGNN, we train a multi-task model to predict distances and hop lengths for two original graph vertices. 

Furthermore, the framework includes a learning-based shortest path search algorithm, called $\textsc{LSearch}$, that can identify shortest paths while exploiting prediction models to prune its search.
The framework employs a vertex skip strategy that uses predicted distances and hop lengths to prune vertices, and it employs a protection strategy to avoid unsafe pruning that would discard relevant vertices.
When dealing with larger graphs, it becomes difficult or impossible to train prediction models on the whole graph. To address this issue, we enable splitting a larger graph into subgraphs so that training can be performed on the subgraphs. We design a hierarchical structure to maintain the models and connections among subgraphs. Moreover, we propose an algorithm, $\textsc{HLSearch}$, to find shortest paths between pairs of vertices in the hierarchical structure of a larger graph.

We report on extensive experiments on five real datasets that offer insight into the workings of the proposed framework.

In summary, we make the following main innovations.
\begin{itemize}
    \item We design a skeleton graph to capture distance and hop information of a generic graph at multiple levels of granularity. (Section~\ref{sec:skeleton_graph})
    \item We propose a Skeleton Graph Neural Network (SGNN) for distance and hop length prediction. (Section~\ref{sec:SGNN})
    \item We propose a learning-based shortest path search method (\textsc{LSearch}) with  pruning strategies enabled by the SGNN.
    We enable splitting a larger graph into subgraphs and provide a hierarchical structure to organize the subgraphs and their corresponding SGNNs. This is leveraged by \textsc{HLSearch} for finding shortest paths in large graphs.
    (Section~\ref{sec:LSPS})
    \item We report on extensive experiments on five real graphs, offering insights and evidence of the efficiency of the proposed framework. (Section~\ref{sec:experiment})
\end{itemize}

In addition, Section~\ref{sec:preliminary} presents the problem and provides a solution overview, Section~\ref{sec:related} reviews related work, and Section~\ref{sec:conclusion} concludes the paper and presents future research directions.

\vspace*{-3pt}
\section{Preliminaries}
\label{sec:preliminary}
Table~\ref{tab:notation} lists the notations used in the paper.
\vspace*{-7pt}
\begin{table}[!htbp]
\centering
\footnotesize
{\setlength\tabcolsep{4pt}
\caption{Notations.}
\label{tab:notation}
\vspace*{-10pt}
\begin{tabular}{c|l}
\toprule
{Symbol} & {Description} \\ \midrule
$v_i$, $V$ & a vertex and the set of vertices \\
$e_{i,j}$, $w_{i,j}$ & the edge and edge weight between vertices $v_i$ and $v_j$ \\ 
$E$, $W$ & the edge set and the weight set \\
$\phi_{s, t}$ & the path from source vertex $v_s$ to target vertex $v_t$ \\
$\delta(\phi_{s, t})$, $\rho(\phi_{s, t})$ & the distance and the hop length of the path $\phi_{s, t}$ \\
$\phi^m$ & $m$-tier skeleton path \\
$\mathcal{L}_i$ & skeleton label of vertex $v_i$ \\
$\mathcal{B}^n_i$ & $n$-hop bucket for $v_i$ \\
\bottomrule
\end{tabular}}
\vspace*{-8pt}
\end{table}

\subsection{Problem Formulation}
\label{ssec:problem}

We consider a generic, undirected graph $G = (V, E, W)$, where each $v_i \in V$ is a vertex, each $e_{i,j} \in E$ is an edge between vertices $v_i$ and $v_j$, and each $w_{i,j} \in W$ is the weight of edge $e_{i,j}$.
Given a source vertex $v_s$ and a target vertex $v_t$, a path from $v_s$ to $v_t$ is a sequence of vertices, denoted as $\phi_{s, t} = (v_0, \ldots, v_{n-1}, v_n)$ where $v_s = v_0$ and $v_t = v_n$. 
In this paper, we also use $\phi_i$ to represent a path when the context is clear.
The \textbf{distance} of a path is defined as $\delta(\phi_{s, t}) = \sum_{(v_i, v_j) \subset \phi_{s,t}}w_{i,j}$. 
The \textbf{hop length} of the path is denoted as $\rho(\phi_{s, t})$ and is the number of edges in the path and thus $n$. 

\noindent\textbf{Shortest Path Search (SPS)}.
Given a graph $G = (V, E, W)$, a source vertex $v_s$ and a target vertex $v_t$, SPS returns the path $\phi_{s, t}$, whose distance $\delta(\phi_{s, t})$ is the shortest among all possible paths $\Phi$ from $v_s$ to $v_t$:
$\nexists \phi'_{s, t} \in \Phi$, $\delta(\phi'_{{s, t}}) < \delta(\phi_{{s, t}})$.

\noindent\textbf{Learning-based SPS} constructs a learning-based model $\mathcal{M}(G)$ of graph $G$ and uses it in a search strategy $\mathcal{S}(\mathcal{M}(G), v_s, v_t)$ that is capable of producing paths that are closely approximate the true shortest path, formally, the objective is
$$
\min_{\mathcal{M}, \mathcal{S}}
\sum\nolimits_{(v_s, v_t) \in V^2} \mathbb{E}[\frac{\lvert\delta(\phi_{s,t}) - \delta(\hat{\phi}_{s,t})\rvert}{\delta(\phi_{s,t})}],
$$
where $\hat{\phi}_{s,t} = \mathcal{S}(\mathcal{M}(G), v_s, v_t)$ is the learning-based path between $v_s$ and $v_t$ and $\phi_{s,t}$ is the true shortest path.

We proceed to provide an overview of our solution to the learning-based SPS problem.

\subsection{Overall Idea and Framework Overview}
\label{ssec:framework}

\vspace*{-10pt}
\begin{figure}[!htbp]
    \centering
    \includegraphics[width=0.85\columnwidth]{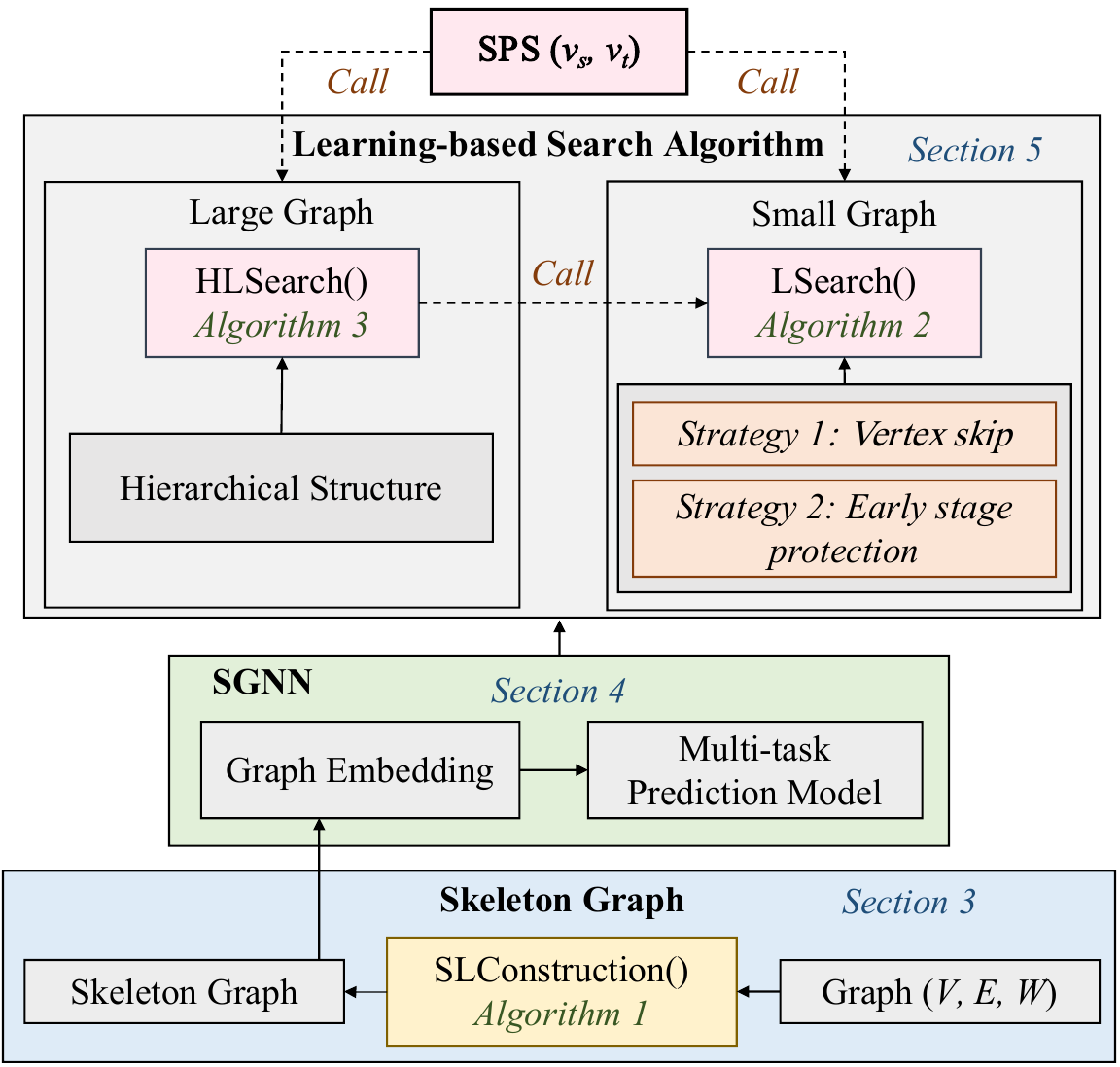}
    \vspace*{-10pt}
    \caption{Overall framework.}
    \label{fig:framework}
\end{figure}
\vspace*{-10pt}

We provide two learning-based shortest path search methods. They avoid some unnecessary expansions during search using predicted path distances and hop lengths. Such predictions are made by a Skeleton Graph Neural Network (SGNN) that abstracts an original graph at multiple levels of granularity and captures all of them in a special graph embedding. Figure~\ref{fig:framework} shows the overall framework.

\begin{figure*}
    \centering
    \subfigure[Original graph]{
        \begin{minipage}[t]{0.63\columnwidth}
            \centering
            \vspace*{-7pt}
            \includegraphics[width=\columnwidth]{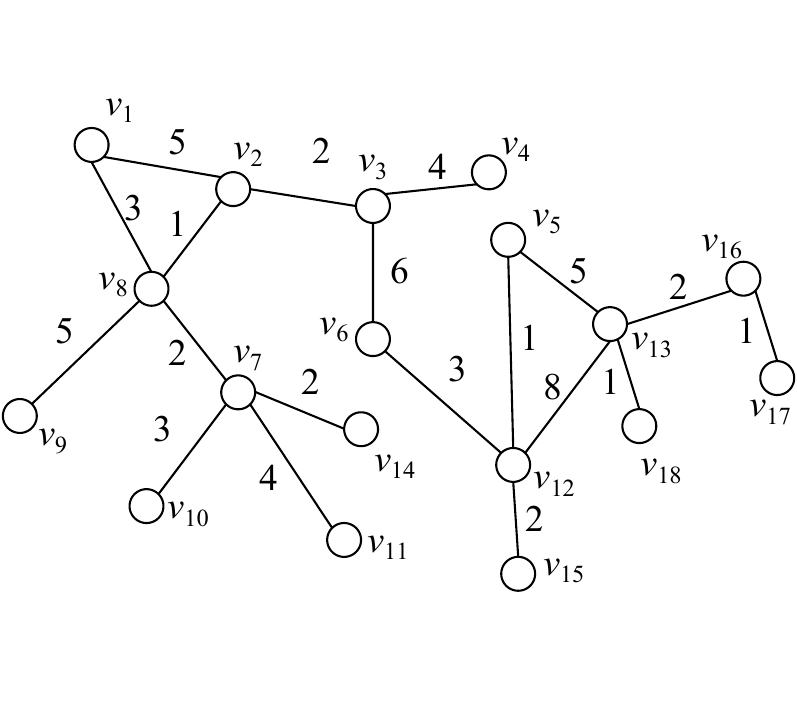}
            \label{fig:example_Graph}
            \vspace*{-10pt}
        \end{minipage}
    }
    \subfigure[0-tier skeleton label $\mathcal{L}^0_1$]{
        \begin{minipage}[t]{0.63\columnwidth}
            \centering
            \vspace*{-7pt}
            \includegraphics[width=\columnwidth]{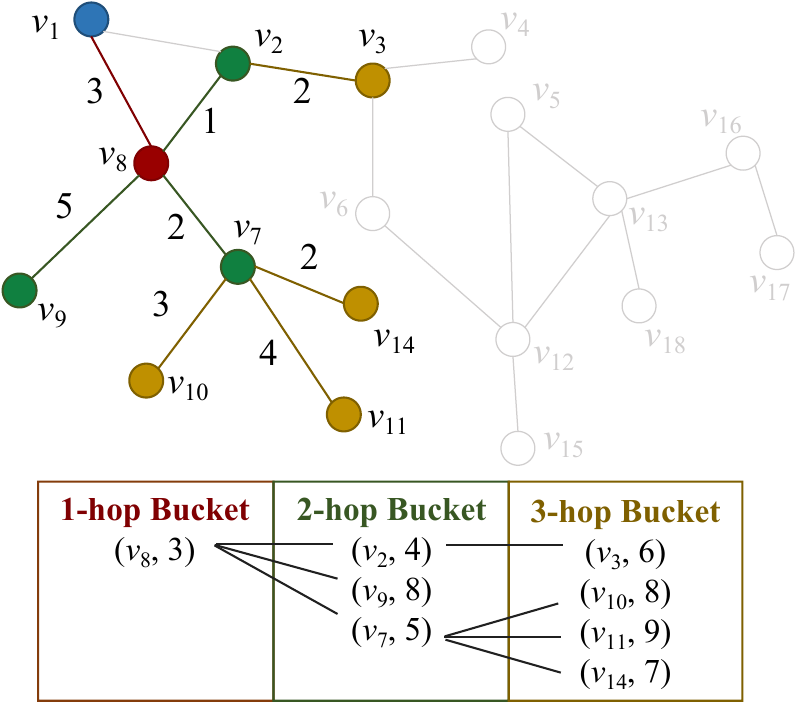}
            \label{fig:0-tier_skeleton}
            \vspace*{-10pt}
        \end{minipage}
    }
    \subfigure[1-tier skeleton label $\mathcal{L}^1_1$]{
        \begin{minipage}[t]{0.63\columnwidth}
            \centering
            \vspace*{-7pt}
            \includegraphics[width=\columnwidth]{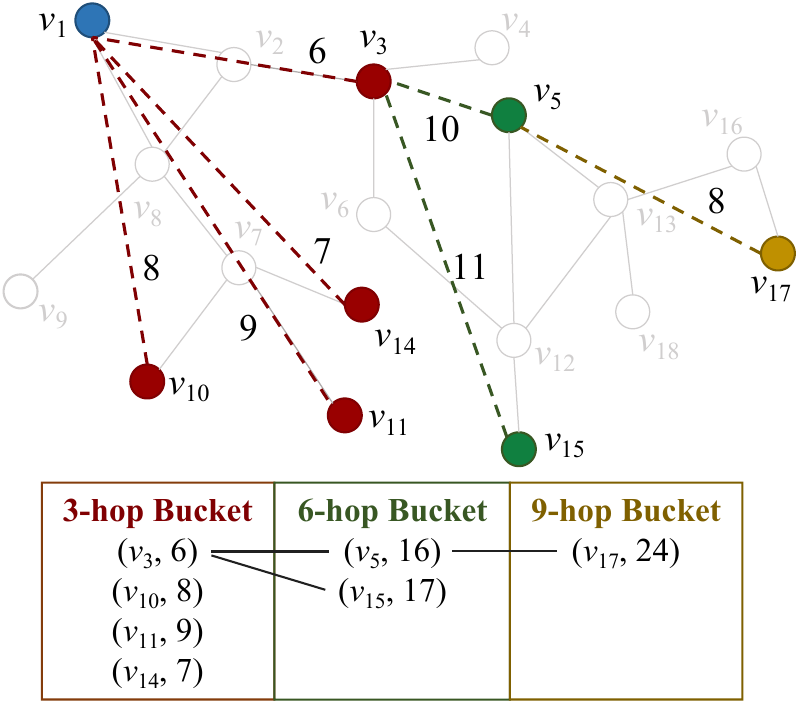}
            \label{fig:1-tier_skeleton}
            \vspace*{-10pt}
        \end{minipage}
    }
    \centering
    \vspace*{-15pt}
    \caption{An example of a skeleton label for $v_1$ ($b = 3$).}
    \label{fig:example_label}
\end{figure*}

In the bottom layer, a skeleton graph is constructed based on the given graph. This will be detailed in Section~\ref{sec:skeleton_graph}. 

The middle layer trains the SGNN, to be detailed in Section~\ref{sec:SGNN}. The SGNN component consists of a graph embedding module and a multi-task prediction model. The model predicts the distance and hop length of the shortest path between two given vertices.
The predictions are used to help accelerate shortest path search.

In the top layer, two learning-based shortest path search algorithms are proposed for small and larger graphs.  
Aiming at small graphs, \textsc{LSearch} exploits with two strategies in its search. The vertex skip strategy uses predicted distances and hop lengths to prune vertices, while the protection strategy guards against unsafe pruning.
As it is difficult to train an SGNN on a larger graph, we design \textsc{HLSearch} for larger graphs. Specifically, we split a larger graph into subgraphs and train an SGNN on each subgraph. Accordingly, we organize the subgraphs and their SGNNs into a hierarchical structure. \textsc{HLSearch} then uses the hierarchical structure.
Algorithms \textsc{LSearch} and \textsc{HLSearch} will be presented in Section~\ref{sec:LSPS}.

\vspace*{-5pt}

\section{Skeleton Graph}
\label{sec:skeleton_graph}

Given a generic graph, shortest path search can benefit from a model that is able to predict shortest distances and hop lengths between pairs of vertices. Such a model can be built on top of appropriate graph embeddings that capture distance- and hop-related information. 
We thus aim to find the vertices reachable from each vertex $v_i$ through shortest distances, and capture such reachable vertices as $v_i$'s labels.
Such vertex labels enable us to reconstruct each vertex's neighbors and to construct skeleton graphs of the original graph, at multiple levels of granularity.
However, if we calculate all vertex-to-vertex distances, the computational and storage overheads will be huge as each vertex's label has size $|V|$. Instead, for each vertex $v_i$, we include in its label only those vertices whose shortest path hop length from $v_i$ is a pre-selected value. We aim to find more vertices topologically close to $v_i$ and select also some vertices relatively topologically far from $v_i$. This way, each vertex can capture information of both near and distant vertices, and the label size is way smaller than $|V|$. Next, we elaborate on how to select labels for each vertex and how to construct the skeleton graph.

\vspace*{-5pt}
\subsection{Skeleton Label}
\label{ssec:skeleton_label}

\begin{definition}[$m$-tier Skeleton Label]
Given a graph $G = (V, E, W)$, a base $b$ that is a positive integer, and a tier number $m$, for each vertex $v_i \in V$, we maintain a skeleton label $\mathcal{L}_i^m$ with $b$ $n$-hop buckets, where $n = k$ $\cdot b^{m}$, $1 \leq k \leq b$. An $n$-hop bucket $\mathcal{B}^n_i$ for $v_i$ stores all vertices to which the shortest path from $v_i$ has $n$ hops. Formally, $\mathcal{B}^n_i = \{v_j \mid  \rho(\phi_{i, j}) = n\}$. 
\end{definition}

Given a vertex $v$, its skeleton label $\mathcal{L}$ is the set of all $m$-tier skeleton labels, i.e., $\mathcal{L} = \{\mathcal{L}^{0}, \mathcal{L}^{1}, \ldots, \mathcal{L}^{M}\}$, where $M$ is the maximum tier number. 
Each skeleton label $\mathcal{L}^{m}$ contains $k$ buckets, and 
two adjacent vertices in adjacent buckets are linked together.

\begin{example}
    Consider $v_1$ as the example in the graph in Figure~\ref{fig:example_Graph}. Figures~\ref{fig:0-tier_skeleton} and~\ref{fig:1-tier_skeleton} illustrate the $0$-tier skeleton label $\mathcal{L}^0_1$ and $1$-tier skeleton label $\mathcal{L}^1_1$ given a base number of $3$.
    Specifically, $\mathcal{L}^0_1$ contains $b = 3$ $n$-hop buckets, with $n \in \{ 1 \cdot 3^0, 2 \cdot 3^0, 3 \cdot 3^0 \} = \{1,2,3\}$.
    Vertex $v_8$ is inserted into the $1$-hop bucket since $v_1$ can reach vertex $v_8$ in $1$ hop with the shortest distance. 
    Although $v_{1}$ can also reach $v_2$ in $1$ hop, the path is not the shortest. Instead, the shortest path from $v_1$ to $v_2$ is $\phi_{1, 2} = (v_1, v_8, v_2)$, and therefore $v_2$ is placed in the $2$-hop bucket.
    Likewise, other vertices having shortest paths from $v_1$ in $2$ hops are $v_7$ and $v_9$; and vertices having $3$-hop shortest paths from $v_1$ include $v_3$, $v_{14}$, $v_{10}$, and $v_{11}$. 
    In the $1$-tier skeleton label $\mathcal{L}^1_1$, there are also ($b=3$) $n$-hop buckets with $n \in \{1 \cdot 3^1, 2 \cdot 3^1, 3 \cdot 3^1 \}$, namely, the $3$-hop, $6$-hop, and $9$-hop buckets.
    We maintain links between adjacent buckets in each $m$-tier skeleton label. Specifically, in $0$-tier skeleton label, the links are the original edges. In Figure~\ref{fig:0-tier_skeleton}, $v_8$ in the $1$-hop bucket links to $v_2$, $v_9$, and $v_7$ in the $2$-hop bucket. In Figure~\ref{fig:1-tier_skeleton}, $v_3$ in the $3$-hop bucket links to $v_5$ and $v_{15}$, because it must pass $v_3$ to reach $v_5$ and $v_{15}$.
\end{example}

\begin{definition}[$m$-tier Skeleton Path]
\label{def:skeleton_path}
Given a base $b$ ($b \geq 1$) and vertex $v_i$'s $m$-tier skeleton label $\mathcal{L}_i^m$, a path $(v_i, \ldots, v_n)$ is an $m$-tier skeleton path of $v_i$
if each vertex $v_j$ ($v_j \neq v_i$) in the path is from a bucket in $\mathcal{L}_i^m$ and the order of these vertices is the same as that of the buckets organized in $\mathcal{L}_i^m$.
The hop length between each two adjacent vertices in such an $m$-tier skeleton path equals $b^m$.
\end{definition}

\begin{example}
In Figure~\ref{fig:0-tier_skeleton}, the path $(v_1, v_8, v_2, v_3)$ is a $0$-tier skeleton path of vertex $v_1$. Its successive vertices on the path are $v_8$, $v_2$, and $v_3$, from the $1$-hop, $2$-hop, and $3$-hop buckets of $v_1$'s $0$-tier skeleton label $\mathcal{L}^0_1$, respectively.
In this path, the hop length between adjacent vertices, i.e., from $v_1$ to $v_8$, from $v_8$ to $v_2$, and from $v_2$ to $v_3$, equals $3^0 = 1$.
Similarly, $(v_1, v_8, v_7, v_{11})$ is a $0$-tier skeleton path of $v_1$.
We can also derive that $(v_1, v_3, v_5, v_{17})$ is a $1$-tier skeleton path of $v_1$ since $v_3$, $v_5$, and $v_{17}$ are from the $3$-hop, $6$-hop, and $9$-hop buckets, respectively. The hop length between each two adjacent vertices equals $3^1 = 3$.
\end{example}

The $m$-tier skeleton label $\mathcal{L}^m$ has the following properties. 
\begin{figure*}[!htbp]
    \centering
    \includegraphics[width=0.95\textwidth]{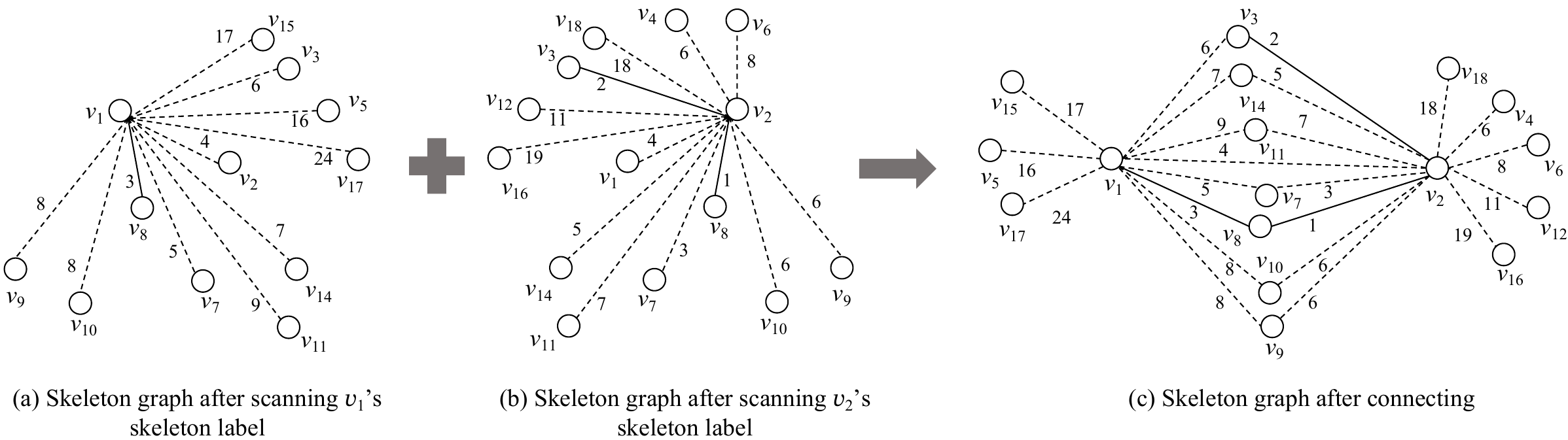}
    \vspace*{-10pt}
    \caption{An example of skeleton graph.}
    \label{fig:example_skeleton_topology}
    \vspace*{-7pt}
\end{figure*}
\vspace*{-7pt}
\begin{property}\label{pro:subset}
    Given a skeleton path $\phi^m = (v_i, \ldots, v_n)$ of vertex $v_i$, $\phi^m$ must be a sub-sequence of $\phi_{i, n}$, the shortest path from $v_i$ to $v_n$. 
\end{property}
The proof of Property~\ref{pro:subset} is straightforward.

\begin{property}\label{pro:skeleton_hop}
    Given a linked vertex pair ($v_i$, $v_j$) in adjacent buckets in an $m$-tier skeleton label ($m > 0$), the $(m-1)$-tier skeleton path $\phi^{m-1}_{i, j}$ must be contained in $v_i$'s $(m-1)$-tier skeleton label $\mathcal{L}^{m-1}_i$.
\end{property}

\begin{proof}
    According to Definition~\ref{def:skeleton_path}, the hop length from $v_i$ to $v_j$ is $b^{m}$. In $v_i$'s $(m - 1)$-tier skeleton label, we have $b$ buckets that each
    stores all vertices to each of which the hop length of the shortest path from $v_i$ is $n$, where $n=$ $k$ $\cdot b^{m - 1}$, $1 \leq k \leq b$. When $k = b$, $n = b^{m}$. The hop length from $v_i$ to $v_j$ is $b^{m}$, and thus we can find it in a bucket in $v_i$'s $(m - 1)$-tier skeleton label.
\end{proof}

\subsection{Skeleton Label Construction}

We construct a skeleton label for each vertex in a graph, which is formalized in Algorithm~\ref{alg:SLConstruction}.

\begin{algorithm}
\small
\caption{\textsc{SLConstruction}($G$, $b$, $m$)} \label{alg:SLConstruction}
    \begin{algorithmic}[1]
        \Statex \textbf{Input:} $G = (V,E,W)$: Graph
        \Statex $b$: The base
        \Statex $m$: The highest tier number
        \Statex \textbf{Output:} $\mathcal{L}$: Skeleton label set
        \State initialize a skeleton label set $\mathcal{L}$
        \For{$v_i$ in $V$}
            \State initialize a skeleton label $\mathcal{L}_i$, a priority queue $\mathcal{Q}$
            \State initialize $\mathit{hop}$[], $\mathit{dist}$[], and $\mathit{prev}$[]
            \For {$v_j$ in $V/v_i$}
                \State $\mathit{hop}[v_j]$ $\gets$ $\infty$, $\mathit{dist}[v_j]$ $\gets$ $\infty$, $\mathit{prev}[v_j]$ $\gets$ $\mathit{null}$
            \EndFor
            \State $\mathit{hop}[v_i]$ $\gets$ $0$, $\mathit{dist}[v_i]$ $\gets$ $0$, $\mathcal{Q}.\mathit{push}(v_i, dist[v_i])$
            \State $\mathit{hop}_{min} \gets \infty$
            \While{$\mathcal{Q}$ is not empty}
                \If{$\mathit{hop}_{min} > b^{m+1}$}
                    $\mathit{break}$
                \EndIf
                \State $v_j$, $\mathit{dist}_j$ $\gets$ $\mathcal{Q}.\mathit{pop}()$
                \If{$\mathit{dist}_j = \infty$} $\mathit{break}$
                \EndIf
                \If{$\mathit{hop}[v_j] > b^{m+1}$} $\mathit{continue}$
                \EndIf
                \For{$v_n$ in adjacent set of $v_j$}
                    \State $\mathit{dist}_n$ $\gets$ $\mathit{dist}_j + w_{j, n}$ 
                    \If{$\mathit{dist}_n < \mathit{dist[v_n]}$}
                        \State $\mathit{hop}[v_n] \gets \mathit{hop}[v_j] + 1$
                        \State $\mathit{dist}[v_n] \gets \mathit{dist}_n$
                        \State $\mathit{prev}[v_n] \gets v_j$
                        \State $\mathcal{Q}.\mathit{update}(v_n, \mathit{dist}[v_n])$
                        \If{$\mathit{hop}[v_n] < \mathit{hop}_{min}$}
                            \State $\mathit{hop}_{min} \gets \mathit{hop}[v_n]$
                        \EndIf
                    \EndIf
                \EndFor
                \If{$\mathit{hop}[v_j] = k \cdot b^{m'}$ }
                    \State $v_{l} \gets \textsc{FindLinkedVertex}(\mathcal{B}^{(k - 1) \cdot b^{m'}}_i, v_j)$ 
                    \State put $(v_j, v_l, \mathit{dist}[v_j]$) into $\mathcal{B}^{k \cdot b^{m'}}_i$
            \EndIf
            \EndWhile
            \State $\mathcal{L}.\mathit{put}(v_i, \mathcal{L}_i)$
        \EndFor
        \State return $\mathcal{L}$
    \end{algorithmic}
\end{algorithm}

First, a skeleton label set $\mathcal{L}$ is initialized to maintain all skeleton labels of vertices in graph $G$ (line~1).
Then, for each vertex $v_i$ in $V$, we construct a skeleton label that contains $b \cdot (m+1)$ buckets (lines~2--26). We first initialize a skeleton label $\mathcal{L}_i$ to store the label information and a priority queue $\mathcal{Q}$ to maintain the information during the expansion (line~3). After that, we initialize and assign three arrays to store the hop number, distance, and previous vertex of the expanded vertex (lines~4--7).
We initialize a hop value to maintain the minimum hop during the search (line~8).
Subsequently, we keep exploring until all labels of $v_i$ are found (lines~9--25). If the minimum hop exceeds $b^{m+1}$, this means all the labels of $v_i$ are found, so the expansion stops (line~10). The vertex with the shortest distance to $v_i$ is popped out from priority queue $\mathcal{Q}$. If the distance is infinity, the expansion for $v_i$ stops because the remaining vertices are unreachable from $v_i$ (lines~11~12). If the hop length is longer than $b^{m + 1}$, it will not further expand from $v_j$ (line~13). Otherwise, we find all adjacent vertices of $v_j$ and keep information of the expansion (lines~14--22). If the hop length of $v_j$ is $k \cdot b^{m'}$, where $1 \leq k \leq b$ and $0 \leq m' \leq m$, we put the information into the corresponding bucket in $\mathcal{L}_i$ (lines~23--25). Specifically, we find the linked previous vertex in the adjacent bucket using function $\textsc{FindLinkedVertex}$ (line~24). After all labels of $v_i$ are found, we put $\mathcal{L}_i$ into $\mathcal{L}$ (line~26). After all skeleton labels are constructed, the overall label information is returned (line~27).

We employ two techniques to accelerate the skeleton label construction. First, for each one-degree vertex, we only keep the 1-hop label in its skeleton label. This reduces the construction cost. Second, we use small values for $b$ and $m$. This also reduces the construction cost, as the search stops when the minimum hop exceeds $b^{m+1}$. 

\subsection{Skeleton Graph}

Given a graph $G = (V,E,W)$ with skeleton label $\mathcal{L}$, we construct a skeleton graph for each vertex using the information of  skeleton label and then combine all the skeleton graphs of vertices.
We use $\mathcal{G} = (V, E', W')$ to denote the combined skeleton graph.

We consider two types of skeleton links. First, \textbf{bucket links} are the same as those between two adjacent buckets in the skeleton label. For example, Figure~\ref{fig:0-tier_skeleton} shows the link $\left \langle v_2, v_3 \right \rangle$ in the $0$-tier skeleton label of $v_1$ and the link $\left \langle v_3, v_5 \right \rangle$ in the $1$-tier skeleton label of $v_1$.
Second, a \textbf{label link} is a link between an original vertex $v_i$ and each vertex in $v_i$'s skeleton label. 
Since each vertex in the bucket can finally link to the original vertex $v_i$, we also keep these links. For example, we will construct a link from $v_1$ to $v_3$ in Figure~\ref{fig:0-tier_skeleton}.

A skeleton graph has more edges, some of which do not exist in the original graph. Further, an original edge may be omitted in the skeleton graph, if the edge weight exceeds the shortest distance between the two pertinent vertices.

To construct the skeleton graph, we scan the labels of each vertex in the graph and construct the two kinds of links. 
Then, we connect  the skeleton graphs of all vertices.
It is easy to show that the bucket link in $v_i$'s label can be found in label links of other vertices according to Property~\ref{pro:skeleton_hop} in Section~\ref{ssec:skeleton_label}. Therefore, we only keep the label links when scanning each vertex's labels. Figure~\ref{fig:example_skeleton_topology} shows an example of the graph after scanning the skeleton labels of $v_1$ and $v_2$. We use a solid line to denote an edge available in the original graph and a dashed line to denote a new edge. Compared to the original graph in Figure~\ref{fig:example_Graph}, more edges are constructed, e.g., $\left \langle v_1, v_9 \right \rangle$. Also, some original edges are assigned with a new weight. For example, the edge weight of $\left \langle v_1, v_2 \right \rangle$ is $5$ in the original graph, but the shortest distance from $v_1$ to $v_2$ is $4$ via $v_8$. Therefore, we assign the weight $4$ to $\left \langle v_1, v_2 \right \rangle$ in the skeleton graph.

\section{Skeleton Graph Neural Network for Distance and Hop Prediction}
\label{sec:SGNN}

\begin{figure*}[!htbp]
    \centering
    \includegraphics[width=0.95\textwidth]{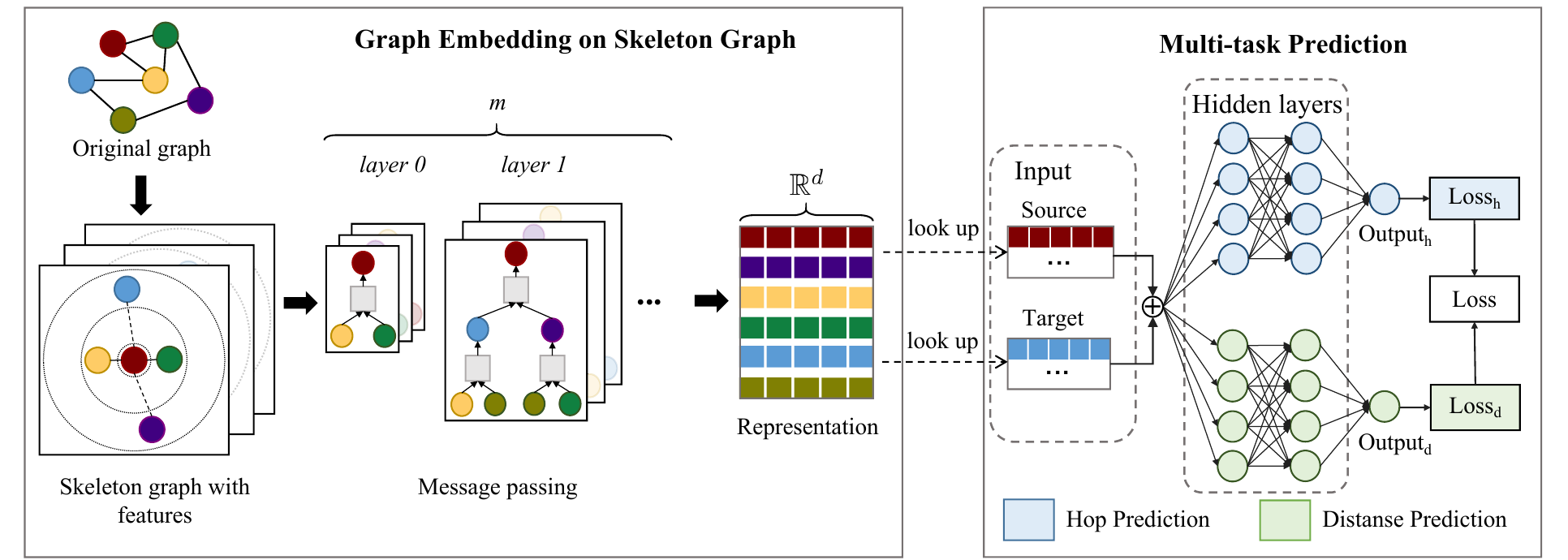}
    \vspace*{-10pt}
    \caption{Skeleton Graph Neural Network (SGNN).}
    \label{fig:sgnn}
    \vspace*{-7pt}
\end{figure*}

With the created skeleton graph, we are able to obtain important features directly related to the distance information of the shortest paths. Next, we exploit the characteristics of the skeleton graph to build a skeleton graph neural network that is able to predict the shortest distance and the hop length between two vertices. 
The overall structure of the model is shown in Figure~\ref{fig:sgnn}. 
First, we generate the features of each vertex based on the statistics over the graph properties and the built skeleton labels. 
Second, we introduce a novel message passing mechanism that takes into account the characteristics of the skeleton graph structure to obtain the embedding of each vertex. 
Third, the output embeddings are fed into a multi-task prediction module to jointly predict the shortest distance and its hop length for given pairs of vertices. 
The whole process is end-to-end such that the prediction tasks can  benefit directly from the characteristics of the skeleton graph.

\subsection{Graph Embedding on Skeleton Graph}

Graph embedding~\cite{perozzi2014deepwalk, tang2015line, grover2016node2vec} maps the vertices and edges of a graph into a continuous low-dimensional space, while preserving the structural information of the graph. Given a graph $G = (V, E, W)$, we use a mapping $f: v_i \mapsto \boldsymbol{y_i} \in \mathbb{R}^d$ to embed each vertex in the graph to a feature vector.
In this way, we represent the proximity information as a $|V| \times d$ matrix, which in turn serves as the features for training a model. 

Usually, graph embedding techniques can be divided into two categories: \emph{shallow} embedding and \emph{deep} embedding. The {shallow} embedding techniques such as  Node2vec~\cite{grover2016node2vec} can only provide a fixed embedding vector for each vertex, i.e., insensitive to the concrete downstream prediction tasks. This is due to the fact that the embedding process and prediction process are usually independent of each other. In contrast, {deep} embedding techniques such as GNN~\cite{4700287} work in an end-to-end mode and they can generate variable embeddings for vertices depending on the concrete tasks. Because of this, {deep} embedding outperforms {shallow} embedding in many graph learning tasks~\cite{zhang2018link,grohe2020word2vec}. Despite the advantages of GNN, it still has some limitations when it is applied to shortest path-related prediction tasks. First, GNN is designed to aggregate features from locally adjacent vertices for downstream prediction while the shortest path may involve vertices that are beyond the scope of local areas. On the other hand, when we try to involve more vertices in by stacking multiple layers of GNN, it will also suffer from the typical over-smoothing issue~\cite{zhou2020towards}. For these reasons, we decide to exploit the characteristics of the skeleton graph and design a neural network that is learned from the skeleton graph for resolving the shortest path-related prediction tasks.

Given a skeleton graph $\mathcal{G}=(V,E',W')$, we generate vertex embeddings $\boldsymbol{z}_i$ for each vertex $v_i$ using a novel message passing method.
Typically, message passing is an iterative process to allow each vertex to aggregate information from its neighbors and update its own representation. The aggregation at each vertex at the $l$-th iteration is defined as follows.
\begin{equation}\label{equation:message_passing}
  \begin{aligned}
  \boldsymbol{h}^{(l)}_i = \sigma(\boldsymbol{W}^{(l)}_{i}\boldsymbol{h}^{(l-1)}_i + \boldsymbol{W}^{(l)}_s\boldsymbol{m}^{(l)}_i + \boldsymbol{b}^{(l)}),
  \end{aligned}
\end{equation}
where $\boldsymbol{h}^{(l)}_i$ is a hidden embedding for the vertex $v_i \in V$ at the $l$-th iteration, and $\boldsymbol{h}^{(0)}_i$ at the initial phase is the input feature of vertex $v_i$; $\sigma$ is the classical activation function $ReLu(\cdot)$; 
$\boldsymbol{m}^{(l)}_i$ is the message aggregated from the information of $v_i$'s neighbors on the graph; and $\boldsymbol{W}^{(l)}_i$ and $\boldsymbol{W}^{(l)}_s$ are two matrices and $\boldsymbol{b}^{(l)}$ is the bias, all of which are learnable.
The vertex feature $\boldsymbol{h}^{(0)}_i$ consists of both graph statistics and label statistical information of the graph properties and the skeleton labels. The former includes the vertex degree and the clustering coefficient, whereas the latter includes the number of vertices in each bucket and the minimal, maximal, and average distances in each bucket. 

\begin{figure}[!htbp]
    \centering
    \includegraphics[width=0.8\columnwidth]{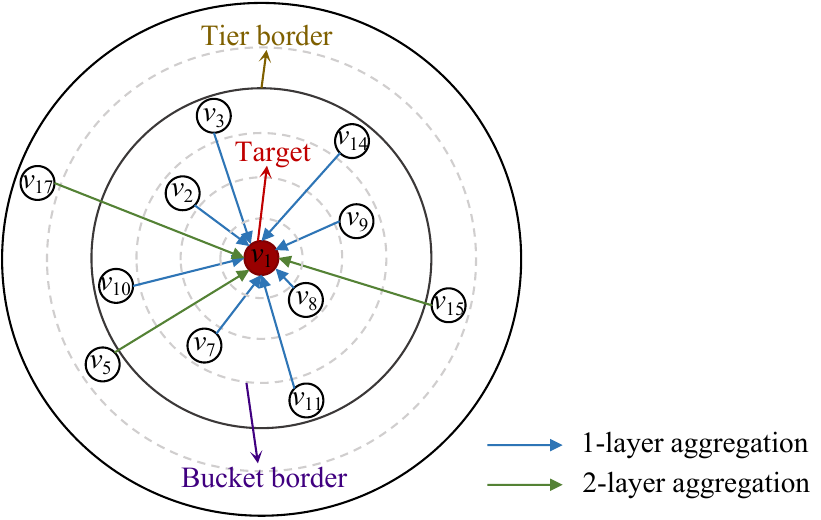}
    \vspace*{-7pt}
    \caption{An example of message passing of our case.}
    \label{fig:mp}
\end{figure}

Instead of using the original graph structure in traditional methods~\cite{hamilton2020graph}, we use the skeleton graph to obtain the neighbors' information for aggregation.
This makes it efficient to consider the information from those topologically far vertices in aggregation.
Specifically, the message $\boldsymbol{m}^{(l)}_i$ at the $l$-th iteration is calculated as
\begin{equation}\label{equation:aggregation}
  \begin{aligned}
    \boldsymbol{m}^{(l)}_i = \sum_{v_j \in \mathcal{N}^{(l)}(v_i)} \frac{\boldsymbol{h}^{(l-1)}_j}{{\sqrt{|\mathcal{N}^{(l)}(v_i)|}}{\sqrt{|\mathcal{N}^{(l)}(v_j)|}}},
  \end{aligned}
\end{equation}
where $\mathcal{N}(v_i)$ is the set of vertices in $v_i$'s $l$-tier skeleton label, i.e., $\mathcal{L}^l_i$. 
For example, for vertex $v_1$ in Figure~\ref{fig:mp}, the set $\mathcal{N}^{(0)}(v_1)$ contains vertices in the label $\mathcal{L}^0_1$, namely $v_8, v_2, v_7, v_9, v_3, v_{14}, v_{10}, v_{11}$, and $\mathcal{N}^{(1)}(v_1) = \{v_5, v_{15}, v_{17}\}$ likewise.
The information of vertices in the neighbor set is aggregated and normalized~\cite{welling2016semi}.
Finally, the vertex presentation $\boldsymbol{h}^{(L)}_i$ at the last iteration $L$ is used as the final embedding of the vertex $v_i$.

Our graph embedding in SGNN distinguishes from GNN in two major aspects.
First, SGNN is learned on the skeleton graph that is reorganized based on the distance and hop information of the original graph. For each vertex $v_i$, we maintain the adjacent or virtual adjacent vertices in different buckets and tiers according to their hop length of the shortest path from $v_i$. Such selected distance- and hop-related information will benefit the distance and hop-length prediction tasks.
Second, during the message passing process, the $l$-tier vertices' information is aggregated to the target vertex at the corresponding $l$-th iteration, while the process only iteratively aggregates the adjacent vertices' information. 
The benefit of our method is twofold. On the one hand, it is able to capture more information by including those vertices with longer hops. On the other hand, it makes a larger proportion for the information from the closer vertices because the information in the earlier layer will not be attenuated during the message passing process.

\vspace*{-5pt}
\vspace*{-10pt}

\subsection{Multi-task Prediction (MTP)}
\label{ssec:mtp}

After the message passing process on the skeleton graph, we obtain the embedding vector of each vertex. We compile these vectors into an embedding dictionary $\{\mathbf{h}_i^{(L)} \mid v_i \in V \}$. As a result, given a source vertex $s$ and a target vertex $t$, their embedding vectors $\mathbf{h}_s$ and $\mathbf{h}_t$ could be found by looking up in the dictionary. 
On this basis, we design a multi-task model to predict the shortest distance and the hop length of the shortest path between two vertices. This is motivated by the fact that the distance and the hop length of the shortest path are related to each other and their predictions based on a multi-task mechanism can benefit each other. 

First, the embedding vectors of the source and target vertices, i.e., $\mathbf{h}_s$ and $\mathbf{h}_t$, are concatenated. The concatenated vector is fed into two separate multilayer perceptrons to predict the distance $\hat{\mathbf{y}}_d$ and the hops $\hat{\mathbf{y}}_h$ of the shortest path, respectively. 
The whole process can be represented by:
$$\hat{\mathbf{y}}_d = MLP_d (\mathbf{h}_s \oplus \mathbf{h}_t)$$
$$\hat{\mathbf{y}}_h = MLP_h (\mathbf{h}_s \oplus \mathbf{h}_t)$$ where $MLP(\cdot)$ represents a multilayer perceptron and $\oplus$ denotes the concatenation operation. 

Furthermore, we can compute the multi-task loss based on the predicted outcomes and their corresponding ground-truth values. Let $\mathbf{y}_d$ and $\mathbf{y}_h$ represent the ground-truth of the shortest distance and hop length respectively. The loss with respect to each of them is denoted as:
\begin{equation}\label{equation:loss_distance}
  \begin{aligned}
  \textit{Loss}_d = MSE( \mathbf{y}_d, \hat{\mathbf{y}}_d) 
  \end{aligned}
\end{equation}
\begin{equation}\label{equation:loss_hop}
  \begin{aligned}
  \textit{Loss}_h = MSE( \mathbf{y}_h, \hat{\mathbf{y}}_h)
  \end{aligned}
\end{equation}
where $MSE(\cdot)$ represents mean squared error function. Based on the separate losses, a multi-task loss is the weighted sum of them:
\begin{equation}\label{equation:loss_distance}
  \begin{aligned}
  \textit{Loss} = \gamma \cdot \textit{Loss}_d + (1-\gamma)\cdot \textit{Loss}_h
  \end{aligned}
\end{equation}
where $\gamma$ is the weight hyperparameter. As our proposed model works in an end-to-end mode, the multi-task loss will be propagated backward through each module of the model, including the embedding layer, to update their parameters. In this way, the graph embedding is optimized.

\section{Learning-Based Search Methods}
\label{sec:LSPS}

Although SGNN is able to predict the shortest distance between two arbitrary vertices, it does not give the shortest path. Nevertheless, we can make use of SGNN to find the shortest paths efficiently and effectively. In this section, we design two learning-based shortest path search methods. 

\subsection{Learning-Based Shortest Path Search}
\label{ssec:LSPS_G_algorithm}

Our search method follows the spirit of a classical shortest path search algorithm. Specifically, it keeps exploring and expanding the adjacent vertices from the source vertex $v_s$ until the target vertex $v_t$ is reached. At each iteration, it determines which adjacent vertex to further expand on the fly. 
Inspired by the A* algorithm~\cite{hart1968formal}, we select vertex $v_i$ that minimizes the following heuristic function:
\begin{equation}
  \begin{aligned}
  \delta'(\phi_{s,t}) = \delta(\phi_{s,i}) + \hat{\delta}(\phi_{i,t}),
  \end{aligned}
\end{equation}
where $\delta(\phi_{s,i})$ is the distance from $v_s$ to $v_i$ calculated so far, and $\hat{\delta}(\phi_{i,t})$ denotes the predicted distance from vertex $v_i$ to the target vertex $v_t$ as per the SGNN model.

We propose a pruning technique to enhance the efficiency of search by skipping certain vertices during expansion. 
Intuitively, if a vertex $v_i$'s current calculated distance from the source $v_s$ is larger than its shortest distance, i.e., $\delta(\phi_{s,i}) - \delta(\phi'_{s,i}) > 0$, we will not further expand on it. Since we do not know the shortest distance, we rely on the SGNN model for prediction. 
However, this prediction can be inaccurate, resulting in errors. To mitigate this, we introduce an error buffer to avoid unsafe prunes.
Furthermore, to enhance the safety of the pruning, we use multiple constraints instead of a single constraint.
For example, for a vertex $v_i$, if the hop length of the current path from the source vertex $v_s$ is not equal to that of the shortest path from the source vertex, i.e., |$\rho(\phi_{s,i})$ - $\rho(\phi'_{s,i})| > 0$, we will terminate the expansion of vertex $v_i$. 
This indicates that the current partial path cannot be part of the shortest path.
Similarly, we also use the predicted hop in the constraint and introduce an error buffer for it. 
We formalize the strategy as follows.


\begin{strategy}[Vertex Skip]\label{strategy:vertex_skip}
When a vertex $v_i$ is popped from the priority queue, it will be skipped without further expansion if it satisfies the following two conditions:
\begin{equation}
  \begin{aligned}
  \delta(\phi_{s,i}) - \hat{\delta}(\phi'_{s,i}) > \alpha \cdot e^d,
  \end{aligned}
\end{equation}
\begin{equation}
  \begin{aligned}
  |\rho(\phi_{s,i}) - \hat{\rho}(\phi'_{s,i})| > \alpha \cdot \ceil{e^h},
  \end{aligned}
\end{equation}
where $\delta(\phi_{s,i})$ and $\rho(\phi_{s,i})$ refer to the current distance and hop length from $v_s$ to $v_i$ during the search, $\hat{\delta}(\phi'_{s,i})$ and $\hat{\rho}(\phi'_{s,i})$ are the predicted distance and hop length of the shortest path from $v_s$ to $v_i$, $e^d$ and $e^h$ are maximal absolute errors\footnote{In the experiments, we use the maximum errors of the distance-prediction and hop-prediction results on the test set.} of the distance-prediction model and the hop-prediction model, respectively, and $\alpha \in [0, 1]$ is a parameter to control the two error buffers.
\end{strategy}

In a real application, if the candidate vertex is skipped at the beginning of the search, it may result in a path much longer than the real shortest path.
To this end, we propose the following strategy to safeguard the candidate vertex during the early stage of the search.

\begin{strategy}[Early Stage Protection]
\label{strategy:protection}
During the expansion, we utilize heuristics to guide the search only if the hop length of the current partial path $\phi_{s,i}$ exceeds a certain parameter $\beta$. Otherwise, we use the shortest distance from the source vertex to the vertex being expanded to guide the search. Formally,
\begin{equation}
  \begin{aligned}
  \delta'(\phi_{s,t}) = 
  \begin{cases}
  \delta(\phi_{s,i}) + \hat{\delta}(\phi_{i,t}), \text{if~} \rho(\phi_{s,i}) > \beta, \\
  \delta(\phi_{s,i}), \text{if~} \rho(\phi_{s,i}) \leq \beta.
  \end{cases}
  \end{aligned}
\end{equation}
Besides, Strategy~\ref{strategy:vertex_skip} is activated only when  $\rho(\phi_{s,i}) > \beta$.  
\end{strategy}

\begin{algorithm}
\small
\caption{\textsc{LSearch}($G$, $\mathcal{M}$, $v_s$, $v_t$)} \label{alg:LSearch}
    \begin{algorithmic}[1]
        \Statex \textbf{Input:} $G = (V,E,W)$: Graph
        \Statex $\mathcal{M}$: trained model
        \Statex $v_s$: Source vertex
        \Statex $v_t$: Target vertex 
        \Statex \textbf{Output:} $\phi_{s, t}$: Shortest path with the  distance from $v_s$ to $v_t$
        \State initialize $\mathit{hop}$[], $\mathit{dist}$[], $\mathit{dist}^t$[], and $\mathit{prev}$[], a priority queue $Q$
        \For {$v_i$ in $V$}
            \State $\mathit{hop}[v_i]$, $\mathit{dist}[v_i]$, $\mathit{dist}^t[v_i]$ $\gets$ $\infty$, $\mathit{prev}[v_i]$ $\gets$ $\mathit{null}$
        \EndFor
        \State $\mathit{hop}[v_s]$, $\mathit{dist}[v_s]$ $\gets$ $0$, \State $\mathit{dist}^t[v_s]$ $\gets$ $\mathcal{M}^d(v_s, v_t)$, $Q.\mathit{push}(v_s, dist^t[v_s])$
        \While{$Q$ is not empty}
            \State $v_i$, $\mathit{dist}_i$ $\gets$ $Q.\mathit{pop}()$, $\mathit{hop}_i \gets \mathit{hop}[v_i]$
            \If{$v_i = v_t$ or $\mathit{dist}_i \geq \mathit{dist}[v_t]$} 
                \State \textbf{return} $\phi_{s, t} \gets \textsc{getPath}(v_s, v_i, \mathit{prev})$
            \EndIf
            \If{$\mathit{hop}$[$v_i$] $> \beta$} \Comment{Strategy~\ref{strategy:protection}}
                \State $\hat{\mathit{dist}_i} \gets \mathcal{M}^d(v_s, v_i)$, $\hat{\mathit{hop}_i} \gets \mathcal{M}^h(v_s, v_i)$
                \If{$\mathit{dist}_i - \hat{\mathit{dist}_i} > \alpha e^d$ and $|\mathit{hop}_i - \hat{\mathit{hop}_i}| > \alpha \ceil{e^h}$}
                \State \textbf{continue}
                \Comment{Strategy~\ref{strategy:vertex_skip}}
                \EndIf
            \EndIf
            \State $\mathit{list} \gets$ adjacent vertices of $v_i$
            \For{$v_j$ in $\mathit{list}$}
                \State $\mathit{dist}_j$ $\gets$ $\mathit{dist}$[$v_i$] + $\mathit{dist}_{i, j}$ 
                \If{$\mathit{dist}_j < \mathit{dist}[v_j]$}
                    \State $\mathit{dist}[v_j] \gets \mathit{dist}_j$
                    \State $\mathit{dist}^t[v_j] \gets \mathit{dist}_j + \mathcal{M}^d(v_j, v_t)$
                    \State $\mathit{prev}[v_j] \gets v_i$
                    \State $\mathit{hop}[v_j] \gets \mathit{hop}_i + 1$
                    \If{$\mathit{hop}[v_j] > \beta$}
                        \Comment{Strategy~\ref{strategy:protection}}
                        \State $Q.\mathit{update}(v_j, \mathit{dist}^t[v_j])$
                    \Else 
                        \State $Q.\mathit{update}(v_j, \mathit{dist}[v_j])$
                    \EndIf
                \EndIf
            \EndFor
        \EndWhile
    \end{algorithmic}
\end{algorithm}

With the heuristic function and the two strategies, our learning based shortest path search method is formalized in Algorithm~{\ref{alg:LSearch}}.
Specifically, it initializes and uses four arrays and a priority queue to maintain the information needed in the search (lines~1--5). 
The array $\textit{dist}^t[]$ maintains the distance from $v_s$ to $v_t$ via each vertex, which influences the vertex ordering in the priority queue $Q$ and guides the expansion. 
The algorithm keeps exploring the neighbors of the expanded vertex until $v_t$ is reached (lines~6--25). 
In each iteration, the vertex $v_i$ with the shortest distance from the source is popped (line~7).
The search will stop if $v_t$ is reached or the popped vertex's distance is larger than the current shortest distance from $v_s$ to $v_t$ (lines~8--9). 
Otherwise, if the hop of $v_i$ is greater than $\beta$ according to Strategy~\ref{strategy:protection} (line 10), we obtain its predicted distance and hops from $v_s$ (line~11). If the predicted values meet the constraints in Strategy~\ref{strategy:vertex_skip}, $v_i$ will be skipped without further expansion (lines~12--13). 
Afterwards, $v_i$'s neighbors are obtained (line~14). For each adjacent vertex $v_j$, we update its shortest distance from $v_s$ (line~15). If the updated shortest distance from $v_s$ is shorter than the currently maintained distance, the information will be updated (lines~17--25). In particular, according to Strategy~\ref{strategy:protection}, if the updated hop length from $v_s$ to $v_j$ is greater than a parameter $\beta$, we put the distance from $v_s$ to $v_t$ via $v_j$ to the queue. Otherwise, we put the distance from $v_s$ to $v_j$ to the queue (lines~22--25).

\subsection{Learning-Based Shortest Path Search on Larger Graphs}
\label{ssec:HLSearch}

When dealing with a larger graph, training an SGNN model on the entire graph can be challenging and time-consuming due to the huge amount of training data. To reduce the training cost, one possible approach is to select only a subset of vertex pairs, which may, however, result in lower model performance.
To address this challenge efficiently and effectively, we propose partitioning the graph into multiple subgraphs, training an SGNN model on each subgraph, and organizing all subgraphs and their corresponding models into a hierarchical structure using a bottom-up approach. This hierarchical structure allows for efficient navigation and retrieval of the subgraphs, enabling fast and accurate querying of the entire graph.
The key steps are presented as follows.
\vspace*{-10pt}
\begin{figure}[!htbp]
    \centering
    \includegraphics[width=0.7\columnwidth]{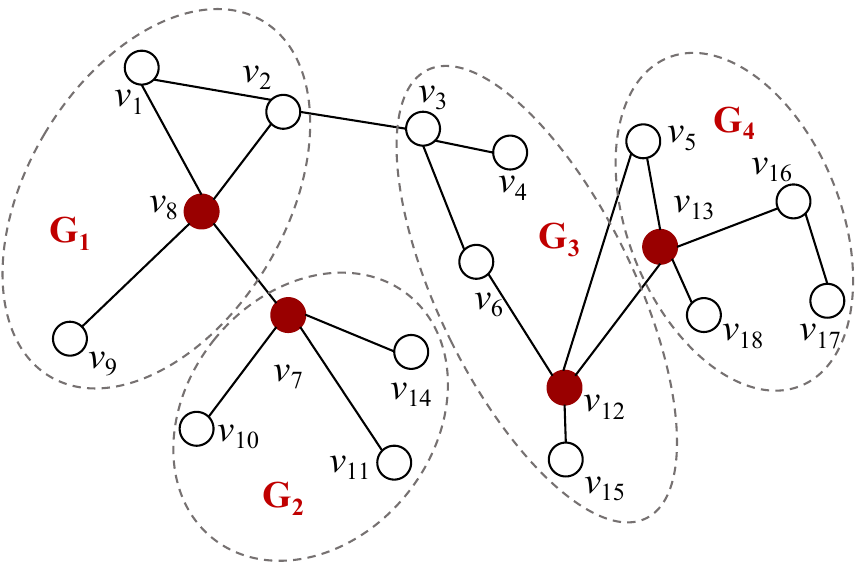}
    \vspace*{-10pt}
    \caption{Graph partitioning.}
    \label{fig:partitioning}
\end{figure}
\vspace*{-10pt}

\smallskip
\noindent\textbf{Graph Partitioning}.
The goal of graph partitioning is to balance the subgraph sizes while minimizing the number of edges connecting different subgraphs. To achieve this goal, we adopt a widely accepted approach~\cite{andreev2004balanced}, 
which starts with generating initial partitions based on some heuristics.
Initially, 
we find $n$ seed vertices with the highest degrees and split the graph into $n$ subgraphs, each of which contains one of the $n$ seed vertices.
Next, we apply iterative refinement to improve the partitions by moving vertices or groups of vertices between partitions until a satisfactory result is obtained.
An example of the resulting graph partitioning is illustrated in Figure~\ref{fig:partitioning}, where $v_8$, $v_7$, $v_{12}$, and $v_{13}$ are the initial seed vertices. We omit the details of graph partitioning since it is beyond the focus of this work. Once the subgraphs are obtained, we construct a skeleton graph for each subgraph and train an SGNN model based on it.

\vspace*{-10pt}
\begin{figure}[!htbp]
    \centering    \includegraphics[width=0.7\columnwidth]{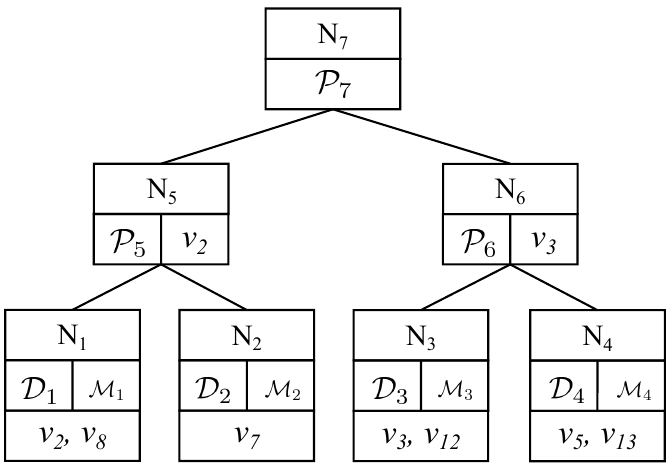}
    \caption{Hierarchical structure.}
    \label{fig:hierarchical}
\end{figure}
\vspace*{-3pt}

\noindent\textbf{Hierarchical Structure}.
To connect the subgraphs, we generate a leaf node for each subgraph and merge adjacent leaf nodes to form non-leaf nodes. These non-leaf nodes are recursively merged into higher-level non-leaf nodes until only one node remains. Each node maintains a set of \textbf{access vertices} that links it to other nodes at the same level.
Each leaf node includes an SGNN and a distance matrix that stores the distance between each vertex in the subgraph and each access vertex in the leaf node. For each non-leaf node, it maintains a distance-path matrix that stores the distance and path between two access vertices of the non-leaf node's children.
Figure~\ref{fig:hierarchical} depicts the hierarchical structure obtained from the graph partitioning shown in Figure~\ref{fig:partitioning}.

\begin{algorithm}
\small    \caption{\textsc{HLSearch}($G$, $\mathcal{H}$, $v_s$, $v_t$)} \label{alg:HLSearch}
    \begin{algorithmic}[1]
        \Statex \textbf{Input:} $G = (V,E,W)$: Graph
        \Statex $\mathcal{H}$: Hierarchical index
        \Statex $v_s$: Source vertex
        \Statex $v_t$: Target vertex 
        \Statex \textbf{Output:} $\phi_{s, t}$: Shortest path with the  distance from $v_s$ to $v_t$
        \State $\mathit{LN}_s \gets \textsc{Leaf}(v_s)$, $\mathit{LN}_t \gets \textsc{Leaf}(v_t)$
        \State $\mathcal{M}_s \gets $ model in $\mathit{LN}_s$, $\mathcal{M}_t \gets $ model in $\mathit{LN}_t$
        \If{$\mathit{LN}_s = \mathit{LN}_t$}
            \State return \textsc{LSearch}($G$, $\mathcal{M}$, $v_s$, $v_t$)
        \EndIf
        \State $\mathit{N}_{LCA} \gets \textsc{LCA}(\mathit{LN}_s, \mathit{LN}_t)$
        \State $\mathit{N}_s \gets$ children of $\mathit{N}_{LCA} \cap$ ancestors of $\mathit{NL}_s$
        \State $\mathit{N}_t \gets$ children of $\mathit{N}_{LCA} \cap$ ancestors of $\mathit{NL}_t$
        \State $V_{cs} \gets$ access vertices in $\mathit{N}_s$,
        $V_{ct} \gets$ access vertices in $\mathit{N}_t$
        \State $V_{ls} \gets$ access vertices in $\mathit{LN}_s$,
        $V_{lt} \gets$ access vertices in $\mathit{LN}_t$
        \State Initialize HashMaps $H_s$ and $H_t$
        \For{$v_i$ in $\mathit{V}_{cs}$}
            \State $\delta(\phi_{s, i}) \gets$ \textsc{getDist}($v_s$, $v_i$)
            \State $H_s.put(v_i, \delta(\phi_{s, i}))$
        \EndFor
        \For{$v_j$ in $\mathit{V}_{ct}$}
            \State $\delta(\phi(_{t, j}) \gets$ \textsc{getDist}($v_t$, $v_j$)
            \State $H_t.put(v_j, \delta(\phi_{t, j}))$
        \EndFor
        \State $\delta_{min} \gets +\infty$
        \For{$v_i$ in $\mathit{V}_{cs}$}
            \For{$v_j$ in $\mathit{V}_{ct}$}
                \State $\delta \gets H_s.get(v_i) + H_t.get(v_j)$
                \If{$\delta < \delta_{min}$}
                    \State $\delta_{min} \gets \delta$
                    \State $v_{cs} \gets v_i$, $v_{ct} \gets v_j$
                \EndIf
            \EndFor
        \EndFor
        \State $v_{ls}, v_{lt} \gets$ \textsc{getLeafAccessVertices}($v_{cs}, v_{ct}$)
        \State $\phi_{s,cs} \gets$ \textsc{LSearch}($G$, $\mathcal{M}_s$, $v_s$, $v_{ls}).append$(\textsc{getPath}($v_{ls},v_{cs}$))
        \State $\phi_{t,ct} \gets$ \textsc{LSearch}($G$, $\mathcal{M}_t$, $v_t$, $v_{lt}).append$(\textsc{getPath}($v_{lt},v_{ct}$))
        \State $\phi_{s,ct} \gets \phi_{s,cs}.append$(\textsc{getPath}($v_{cs},v_{ct}$))
        \State $\phi_{s,t} \gets \phi_{s,ct}.append$($\phi_{t,ct}.reverse()$)
        \State return $\phi_{s,t}$
    \end{algorithmic}
\end{algorithm}

\smallskip
\noindent\textbf{Shortst Path Search}. Algorithm~\ref{alg:HLSearch} finds the shortest path on the hierarchical structure of a larger graph. 
It first finds $v_s$'s and $v_t$'s leaf nodes $LN_s$ and $LN_t$ using a function $\textsc{Leaf}()$ (line~1). Next, it gets the prediction models $\mathcal{M}_s$ and $\mathcal{M}_t$ in the two leaf nodes.
If $v_s$ and $v_t$ are in the same leaf node, it calls $\textsc{LSearch}$ and return the shortest path (lines~3--4).
Otherwise, it retrieve the related nodes and access vertices.
Specifically, It finds the lowest common ancestor $N_{LCA}$ (line~5), and then it gets $N_{LCA}$'s children $N_s$ and $N_t$ that are $LN_s$'s and $LN_t$'s ancestors, respectively (lines~6--7).
Afterwards, it gets the access vertices from nodes $N_s$, $N_t$, $LN_s$ and $LN_t$ (lines~8--9).
Two HashMaps are initialized to maintain the distance during the retrieving (lines~10).
For each access vertex $v_i$ in $V_{cs}$, we get the distance from $v_s$ to $v_i$ and maintain the value in $H_s$ (lines~11--13).
Similarly, we maintain all the distances from $v_t$ to the access vertices in $V_{ct}$ (lines~14--16).
Furthermore, we get the corresponding access vertices that connect $v_s$ and $v_t$ with the shortest distance (lines~17--24).
Since no path information is stored in the leaf node, the algorithm calls $\textsc{LSearch}$ to finds the path from $v_s$ to the access vertex $v_{ls}$ in leaf node $LN_s$, and it appends the path from $v_{ls}$ to the access vertex $v_{cs}$ in $N_s$ (line~25).
Likewise, we get the shortest path from $v_t$ to the corresponding access vertex $v_{ct}$ in $N_t$ (line~26).
Then, we get the path from $v_{cs}$ to $v_{ct}$ and appends it to $\phi_{s,cs}$ (line~27).
Afterwards, we reverse the path $\phi_{t,ct}$ and append it to $\phi_{s,ct}$ (line~28).
Finally, the shortest path from $v_s$ and $v_t$ is returned (line~29).

\section{Experiments}
\label{sec:experiment}

\subsection{Overall Experimental Settings}
\label{ssec:setting}
All index construction and search algorithms are implemented in Java and run on a PC with an Apple M1 chip and 16 GB memory.  All models are coded in Python 3.8 and run on a Linux server with 3.2 GHz Intel Core i9 CPU and NVIDIA Geforce P8 GPU with 24.5 GB memory. All neural network models are implemented using PyTorch 1.6 and trained on the GPU. 

\noindent\textbf{Datasets.} We evaluate our proposed methods on the following five datasets. Their statistics are given in Table~\ref{tab:dataset}.
\begin{itemize}
    \item \textbf{Brain\footnote{\url{https://neurodata.io/project/connectomes/}}:} A brain neuronetwork.
    \item \textbf{Bio\footnote{\url{https://networkrepository.com/bio-SC-LC.php}}:} A biological network which describes gene functional associations. 
    \item \textbf{Web\footnote{\url{https://networkrepository.com/web-EPA.php}}:} A web network which describes page links.
    \item \textbf{Power\footnote{\url{https://networkrepository.com/inf-power.php}}:} A network which represents the topology of the Western States Power Grid of the United States.
    \item \textbf{Road-NA\footnote{\url{https://www.cs.utah.edu/~lifeifei/SpatialDataset.htm}}:} The road network of North America.
\end{itemize}
%
\begin{table}[!htbp]
    \caption{Datasets statistics.}\label{tab:dataset}
    \vspace*{-10pt}
    \small
    \centering
    \begin{tabular}{|l|r|r|r|r|}
    \hline
    \textbf{Dataset} & \textbf{\#Vertices} & \textbf{\#Edges}  & \makecell[c]{\textbf{Maximum} \\ \textbf{Degree}}  & \makecell[c]{\textbf{Average} \\ \textbf{Degree}}                                      
    \\ \hline 
    Brain      & 503 & 24,442 & 497 & 91                                          \\ \hline
    Bio      & 1,999 & 20,448 & 167 & 20                                          \\ \hline
    Web      & 4,253 & 8,897 & 175 & 4                                          \\ \hline
    Power      & 4,941 & 6,594 & 19 & 2                                          \\ \hline
    Road-NA      & 175,813 & 179,102 & 14 & 2                                          \\ \hline
    
    \end{tabular}
\end{table}
\noindent\textbf{Tasks.} We conduct extensive experiments to answer and analyse the following research questions (RQs):
\begin{itemize}
    \item \textbf{RQ1:} How does the the Skeleton Graph Neural Network (SGNN) perform on distance-prediction and hop-prediction tasks on different graphs?
    \item \textbf{RQ2:} How do the learning-based shortest path search methods \textsc{LSearch} and \textsc{HLSearch} perform on different graphs?
\end{itemize}

\noindent\textbf{Metrics.} The metrics for each task is listed in Table~\ref{tab:metrics}.
For SGNN, we evaluate its training time and model size. To evaluate the effectiveness of SGNN, we use metrics the Mean Absolute Percentage Error ($MAPE$) and Root Mean Square Error ($RMSE$):
\begin{equation}
  \begin{aligned}
  MAPE = \frac{1}{n}\sum_{i=1}^n \lvert\frac{\mathbf{y}_i- \hat{\mathbf{y}}_i}{\mathbf{y}_i}\rvert
  \end{aligned}
\end{equation}
\begin{equation}
  \begin{aligned}
  RMSE = \sqrt{\frac{\sum_{i=1}^n \lVert\mathbf{y}_i- \hat{\mathbf{y}}_i\rVert^2}{n}}
  \end{aligned}
\end{equation}
where $n$ is the size of the test set, $\mathbf{y_i}$ is the ground-truth value, and $\hat{\mathbf{y}}_i$ is predicted value. In this work, we evaluate $MAPE$ and $RMSE$ for distance prediction and hop prediction, which are denoted as $MAPE_d$, $MAPE_h$, $RMSE_d$, and $RMSE_h$, respectively.

For \textsc{LSearch} and \textsc{HLSearch}, we measure the query time and the memory use during the search to evaluate their efficiency. We use the accuracy ($Acc$) and hit rate ($Hit$) to evaluate their effectiveness:

\begin{equation}
\label{equ:acc}
  \begin{aligned}
  Acc = \frac{1}{n}\sum_{i=1}^n (1 - \lvert\frac{\delta(\phi_i)- \delta(\hat{\phi}_i)}{\delta(\phi_i)}\rvert)
  \end{aligned}
\end{equation}

\begin{equation}
\label{equ:hit}
  \begin{aligned}
  Hit = \frac{\sum_{i=1}^n \mathcal{S}(\phi_i, \hat{\phi}_i)}{n}
  \end{aligned}
\end{equation}

Above, $n$ is the number of query instances, $\delta(\phi_i)$ is the ground-truth distance and $\delta(\hat{\phi}_i)$ is the calculated distance, $\phi_i$ is the ground-truth path and $\hat{\phi}_i$ is the found path, and function $\mathcal{S}(\phi_i, \hat{\phi}_i)$ returns 1 if $\phi_i = \hat{\phi}_i$ and 0 otherwise.

\begin{table}[!htbp]
\vspace*{-10pt}
\centering
\small
\caption{Metrics.}
\vspace*{-10pt}
\label{tab:metrics}
\begin{tabular}{|c|cc|}
\hline
\multirow{2}{*}{\textbf{Task}} & \multicolumn{2}{c|}{\textbf{Metrics}}                                                                                                                                                             \\ \cline{2-3} 
                        & \multicolumn{1}{c|}{\textbf{Efficiency}}                                                                         & \textbf{Effectiveness}                                                                  \\ \hline
SGNN                    & \multicolumn{1}{c|}{Training time, Size}                                                                & MAPE, RMSE                                                                     \\ \hline
\textsc{LSearch}                 & \multicolumn{1}{c|}{\multirow{2}{*}{\begin{tabular}[c]{@{}c@{}}Query time, \\ Memory use\end{tabular}}} & \multirow{2}{*}{\begin{tabular}[c]{@{}c@{}}Accuracy, \\ Hit rate\end{tabular}} \\ \cline{1-1}
\textsc{HLSearch}                & \multicolumn{1}{c|}{}                                                                                   &                                                                                \\ \hline
\end{tabular}
\end{table}
\vspace*{-10pt}

\noindent\textbf{Baselines.}
We compare our methods to the baseline methods. 
Specifically, we compare SGNN to GNN+MTP and Node2Vec+MTP, where MTP is our proposed multi-task prediction model (Section~\ref{ssec:mtp}).
\begin{itemize}
    \item GNN: We use Graph Neural Network (GNN)~\cite{4700287} to get the representation of each vertex as a baseline. GNNs use pairwise message passing, such that graph vertices iteratively update their representations by exchanging information with their neighbors. The target tasks are same with SGNN, i.e., distance-prediction task and hop-prediction task.
    \item Node2Vec: Another baseline is Node2Vec~\cite{grover2016node2vec}, which learns representations for vertices in a graph by optimizing a neighborhood preserving objective. The target tasks are same with SGNN.
\end{itemize}

For training of neural network models, we set the learning rate to 0.01, the batch size to 10000,  the training epochs to 200, and the embedding size to 32.
The Adam optimizer is used for optimization. All neural networks are tuned to optimal for evaluations.

We implement the following baseline methods for path search:
\begin{itemize}
    \item $\textsc{Dijkstra}$: $\textsc{Dijkstra}$ algorithm~\cite{dijkstra1959note} uses a min-priority queue for storing and querying partial solutions sorted by distance from the start. It also maintains the path information during the search, such that we can find the shortest path.
    \item $\textsc{Landmark}$: We select a set of vertices which are called landmarks~\cite{qiao2012approximate}. For each landmark vertex, we compute its shortest distances to all vertices and store them in a matrix.
    \item $\textsc{HSearch}$: It is a search method on top of the hierarchical structure, which however uses no learning-based strategy. 
\end{itemize}
Specifically, we compare $\textsc{LSearch}$ with $\textsc{Dijkstra}$ and $\textsc{Landmark}$ method, and $\textsc{HLSearch}$ with $\textsc{Dijkstra}$ and $\textsc{HSearch}$.

\subsection{Evaluation of SGNN}
\subsubsection{Construction of the Skeleton Graph}
\label{ssec:construction_sg}

We evaluate the construction time and size of the skeleton graph on different datasets. We set the base $b$ and the tier number $m$ according to each dataset's characteristics. For example, the average and maximum degrees of Power and Road-NA datasets are relatively low, and the max hop of the path over the graph is very long. In this case, we set a larger base value to maintain more edges that can connect two vertices topologically far apart.
The results are given in Table~\ref{tab:overall_construction}. The longest construction time is slightly over 7 minutes and the skeleton graph size is less than 52 MB, all acceptable for offline preprocessing.

\vspace*{-10pt}
\begin{table}[!htbp]
\caption{Performance of skeleton graph construction.}\label{tab:overall_construction}
\vspace*{-10pt}
\small
\begin{tabular}{|l|l|l|l|l|}
\hline
\textbf{Dataset} & \textbf{Time} (s) & \textbf{Size} (MB) & \textbf{b} & \textbf{m}  \\ \hline 
Brain & 7 & 2 & 2 & 2 \\ \hline
Bio & 104 & 23.4 & 2 & 2 \\ \hline
Web & 294 & 51.6 & 2 & 2 \\ \hline
Power & 422 & 9.5 & 3 & 2 \\ \hline
Road-NA & 192 & 29.5 & 3 & 2 \\ \hline
\end{tabular}
\end{table}
\vspace*{-10pt}

\subsubsection{Quality of Models}
To evaluate the quality of SGNN, we measure $MAPE$ and $RMSE$, and compare our proposed model to classical GNN~\cite{hamilton2017inductive} and Node2Vec~\cite{grover2016node2vec}, which are adapted to distance-prediction and hop-prediction tasks. The results of $MAPE$ and $RMSE$ are shown in Tables~\ref{tab:mape} and~\ref{tab:rmse}, respectively.
We observe that SGNN outperforms the two baselines on most datasets. That is because SGNN is learned on the skeleton graph which can capture more distance-related information.
Besides, during the message passing process, SGNN can aggregate more information from farther vertices while guaranteeing that the closer vertices' information has a larger proportion.
\vspace*{-10pt}
\begin{table}[!htbp]
\caption{MAPE of distance prediction and hop prediction.}\label{tab:mape}
\vspace*{-10pt}
\footnotesize
\begin{tabular}{|l|ccc|ccc|}
\hline
\multicolumn{1}{|c|}{\multirow{2}{*}{Dataset}} & \multicolumn{3}{c|}{$MAPE_d$}                                                                            & \multicolumn{3}{c|}{$MAPE_h$}                                                                           \\ \cline{2-7} 
\multicolumn{1}{|c|}{}                         & \multicolumn{1}{c|}{GNN} & \multicolumn{1}{c|}{Node2Vec} & SGNN                              & \multicolumn{1}{c|}{GNN} & \multicolumn{1}{c|}{Node2Vec} & SGNN                             \\ \hline
Brain                                          & \multicolumn{1}{c|}{10.20\%}   & \multicolumn{1}{c|}{10.19\%}        & \textbf{6.13\%}  & \multicolumn{1}{c|}{5.97\%}    & \multicolumn{1}{c|}{7.86\%}         & \textbf{4.07\%} \\ \hline
Bio                                            & \multicolumn{1}{c|}{10.59\%}   & \multicolumn{1}{c|}{12.76\%}        & \textbf{10.24\%} & \multicolumn{1}{c|}{\textbf{
6.25\%}}  & \multicolumn{1}{c|}{14.21\%}        & 10.08\%                          \\ \hline
Web                                            & \multicolumn{1}{c|}{12.71\%}   & \multicolumn{1}{c|}{17.96\%}        & \textbf{8.62\%}  & \multicolumn{1}{c|}{10.42\%}   & \multicolumn{1}{c|}{14.70\%}        & \textbf{7.21\%} \\ \hline
Power                                          & \multicolumn{1}{c|}{18.78\%}   & \multicolumn{1}{c|}{27.77\%}        & \textbf{8.03\%}  & \multicolumn{1}{c|}{17.74\%}   & \multicolumn{1}{c|}{26.38\%}        & \textbf{7.72\%} \\ \hline
Road-NA                                        & \multicolumn{1}{c|}{16.60\%}   & \multicolumn{1}{c|}{42.00\%}        & \textbf{6.00\%}  & \multicolumn{1}{c|}{18.20\%}   & \multicolumn{1}{c|}{43.40\%}        & \textbf{7.40\%} \\ \hline
\end{tabular}
\end{table}
\vspace*{-10pt}
\vspace*{-10pt}
\begin{table}[!htbp]
\footnotesize
\caption{RMSE of distance prediction and hop prediction.}\label{tab:rmse}
\vspace*{-10pt}
\begin{tabular}{|l|ccc|ccc|}
\hline
\multicolumn{1}{|c|}{\multirow{2}{*}{Dataset}} & \multicolumn{3}{c|}{$RMSE_d$}                                           & \multicolumn{3}{c|}{$RMSE_h$}                                              \\ \cline{2-7} 
\multicolumn{1}{|c|}{}                         & \multicolumn{1}{c|}{GNN} & \multicolumn{1}{c|}{Node2Vec} & SGNN  & \multicolumn{1}{c|}{GNN} & \multicolumn{1}{c|}{Node2Vec} & SGNN  \\ \hline
Brain                                          & \multicolumn{1}{c|}{0.33}      & \multicolumn{1}{c|}{0.40}            & \textbf{0.24}  & \multicolumn{1}{c|}{0.30}       & \multicolumn{1}{c|}{0.38}           & \textbf{0.23}  \\ \hline
Bio                                            & \multicolumn{1}{c|}{\textbf{0.35}}      & \multicolumn{1}{c|}{1.31}           & 1.06  & \multicolumn{1}{c|}{\textbf{0.30}}       & \multicolumn{1}{c|}{0.77}           & 0.55  \\ \hline
Web                                            & \multicolumn{1}{c|}{0.76}      & \multicolumn{1}{c|}{0.99}           & \textbf{0.54} & \multicolumn{1}{c|}{0.76}      & \multicolumn{1}{c|}{0.99}           & \textbf{0.55} \\ \hline
Power                                          & \multicolumn{1}{c|}{4.84}      & \multicolumn{1}{c|}{6.48}           & \textbf{2.16}  & \multicolumn{1}{c|}{4.83}      & \multicolumn{1}{c|}{6.48}           & \textbf{2.19}  \\ \hline
Road-NA                                        & \multicolumn{1}{c|}{261.97}    & \multicolumn{1}{c|}{611.34}         & \textbf{111.5} & \multicolumn{1}{c|}{65.66}     & \multicolumn{1}{c|}{153.78}         & \textbf{23.92} \\ \hline
\end{tabular}
\end{table}
\vspace*{-10pt}
\subsubsection{Training Time and Model Size}
Referring to Table~\ref{tab:time_size}, we record the consumed training time until models converge and the size of each model. Overall, {Node2Vec} is the fastest due to its relatively simple training process. Besides, {SGNN} and {GNN} take comparable time to train models in small datasets 
whereas {SGNN} is much faster than that of {GNN} in larger datasets such as {road-NA}. This is because that {SGNN} can exploit the benefit of {skeleton} graph the most on larger graphs to help the model converge to the optimal results. Regarding the size of model, overall {SGNN} is the most efficient and by contrast {Node2Vec} is more memory-consuming. This is relevant to the structure of models and {Node2Vec} calls for more space to store variables than that of both {GNN} and {SGNN}.
These results indicate that our proposed {SGNN} is both time- and space-efficient.
\vspace*{-10pt}
\begin{table}[!htbp]
\footnotesize
\caption{Training time and model size.}\label{tab:time_size}
\vspace*{-10pt}
\begin{tabular}{|l|ccc|ccc|}
\hline
\multicolumn{1}{|c|}{\multirow{2}{*}{Dataset}} & \multicolumn{3}{c|}{Training Time (mins)}                         & \multicolumn{3}{c|}{Size (MB)}                                                \\ \cline{2-7} 
\multicolumn{1}{|c|}{}                         & \multicolumn{1}{c|}{GNN} & \multicolumn{1}{c|}{Node2Vec} & SGNN       & \multicolumn{1}{c|}{GNN} & \multicolumn{1}{c|}{Node2Vec} & SGNN   \\ \hline
Brain                                          & \multicolumn{1}{c|}{16.5} & \multicolumn{1}{c|}{0.08}        & 10.16 & \multicolumn{1}{c|}{0.0196}    & \multicolumn{1}{c|}{0.126}          & 0.0196 \\ \hline
Bio                                            & \multicolumn{1}{c|}{14.9} & \multicolumn{1}{c|}{17.31}     & 19.77 & \multicolumn{1}{c|}{0.0196}    & \multicolumn{1}{c|}{0.126}          & 0.021  \\ \hline
Web                                            & \multicolumn{1}{c|}{0.94}    & \multicolumn{1}{c|}{0.92}        & 17.48   & \multicolumn{1}{c|}{0.025}     & \multicolumn{1}{c|}{0.126}          & 0.02   \\ \hline
Power                                          & \multicolumn{1}{c|}{0.48}    & \multicolumn{1}{c|}{0.92}        & 2.01    & \multicolumn{1}{c|}{0.044}     & \multicolumn{1}{c|}{0.126}          & 0.032  \\ \hline
Road-NA                                        & \multicolumn{1}{c|}{674.41} & \multicolumn{1}{c|}{129.09}      & 295.3   & \multicolumn{1}{c|}{0.044}     & \multicolumn{1}{c|}{0.125}          & 0.032  \\ \hline
\end{tabular}
\end{table}
\vspace*{-10pt}

\subsection{Evaluation of Query Methods}

\subsubsection{Shortest Path Query on Small Graph}

We evaluate the efficiency and effectiveness of $\textsc{LSearch}$ and the baseline methods on the four small datasets. 
We randomly generate 100 query instances, i.e., source and target pairs, and run each method 10 times for each query instance. 

\textbf{Effectiveness.} As both $\textsc{Landmark}$ and $\textsc{LSearch}$ may return approximate results, we measure their effectiveness by groundtruth shortest path hit rate (Equ.~\ref{equ:hit}) shortest path distance accuracy (Equ.~\ref{equ:acc}).
The results are reported in Table~\ref{tab:effectiveness_SPS}.
As we can see, $\textsc{LSearch}$ significantly outperforms $\textsc{Landmark}$ on both metrics. In particular, $\textsc{LSearch}$'s hit rate is around $90$\% , which means it is able to return the groundtruth shortest path in most cases. Its shortest path distance accuracy is around $99$\%, indicating the underlying SGNN works highly effectively.

\vspace*{-10pt}
\begin{table}[!htbp]
\small
\caption{Effectiveness of shortest path search methods.}\label{tab:effectiveness_SPS}
\vspace*{-10pt}
\begin{tabular}{|l|ll|ll|}
\hline
\multicolumn{1}{|c|}{\multirow{2}{*}{Dataset}} & \multicolumn{2}{c|}{Hit Rate (\%)}       & \multicolumn{2}{c|}{Accuracy (\%)}       \\ \cline{2-5} 
\multicolumn{1}{|c|}{}                         & \multicolumn{1}{l|}{Landmark} & LSearch & \multicolumn{1}{l|}{Landmark} & LSearch \\ \hline
Brain                                          & \multicolumn{1}{l|}{70}       & 90       & \multicolumn{1}{l|}{77.5}     & 99.5     \\ \hline
Bio                                            & \multicolumn{1}{l|}{13}       & 87       & \multicolumn{1}{l|}{70.97}    & 98.12    \\ \hline
Web                                            & \multicolumn{1}{l|}{9}        & 99       & \multicolumn{1}{l|}{52.36}    & 99.5     \\ \hline
Power                                          & \multicolumn{1}{l|}{19}       & 84       & \multicolumn{1}{l|}{82.76}    & 98.63    \\ \hline
\end{tabular}
\end{table}
\vspace*{-10pt}

\textbf{Efficiency.} Table~\ref{tab:efficiency_SPS} presents the average query time and memory use of each method. 
The results show that $\textsc{LSearch}$ clearly outperforms $\textsc{Dijkstra}$ in terms of query time. In particular, $\textsc{LSearch}$ is more than 10 times faster than $\textsc{Dijkstra}$ on Power dataset. 
With the learning-based pruning strategy (Strategy~\ref{strategy:vertex_skip} in Section~\ref{ssec:LSPS_G_algorithm}), $\textsc{LSearch}$ prunes some unnecessary expansion, which makes the path search more efficiently.
For memory use, we measure the information storage, including the graph information and pre-computed distances and paths, and the memory used during the search process.
Although $\textsc{Landmark}$  incurs shorter query time than $\textsc{Dijkstra}$ and $\textsc{LSearch}$, it uses much more memory, not only on the information storage but also on the memory used by search.

\begin{table*}[!htbp]
\small
\caption{Efficiency of shortest path search methods.}\label{tab:efficiency_SPS}
\vspace*{-10pt}
\begin{tabular}{|l|lll|llllll|}
\hline
\multicolumn{1}{|c|}{\multirow{3}{*}{Dataset}} & \multicolumn{3}{c|}{\multirow{2}{*}{Query Time (ms)}}                          & \multicolumn{6}{c|}{Memory Use (MB)}                                                                                                                                         \\ \cline{5-10} 
\multicolumn{1}{|c|}{}                         & \multicolumn{3}{c|}{}                                                    & \multicolumn{3}{c|}{Information}                                                               & \multicolumn{3}{c|}{Search}                                              \\ \cline{2-10} 
\multicolumn{1}{|c|}{}                         & \multicolumn{1}{l|}{Dijkstra} & \multicolumn{1}{l|}{Landmark} & LSearch & \multicolumn{1}{l|}{Dijkstra} & \multicolumn{1}{l|}{Landmark} & \multicolumn{1}{l|}{LSearch} & \multicolumn{1}{l|}{Dijkstra} & \multicolumn{1}{l|}{Landmark} & LSearch \\ \hline
Brain                                          & \multicolumn{1}{l|}{14}       & \multicolumn{1}{l|}{5}        & 9        & \multicolumn{1}{l|}{26.79}    & \multicolumn{1}{l|}{50.62}    & \multicolumn{1}{l|}{26.81}    & \multicolumn{1}{l|}{26.02}    & \multicolumn{1}{l|}{26.13}    & 25.34    \\ \hline
Bio                                            & \multicolumn{1}{l|}{91}       & \multicolumn{1}{l|}{7}        & 25       & \multicolumn{1}{l|}{17.91}    & \multicolumn{1}{l|}{25.71}    & \multicolumn{1}{l|}{17.93}    & \multicolumn{1}{l|}{10.372}   & \multicolumn{1}{l|}{10.38}    & 10.341   \\ \hline
Web                                            & \multicolumn{1}{l|}{263}      & \multicolumn{1}{l|}{7}        & 40       & \multicolumn{1}{l|}{38.1}     & \multicolumn{1}{l|}{86.58}    & \multicolumn{1}{l|}{38.12}    & \multicolumn{1}{l|}{20.91}    & \multicolumn{1}{l|}{20.82}    & 20.64    \\ \hline
Power                                          & \multicolumn{1}{l|}{280}      & \multicolumn{1}{l|}{8}        & 22       & \multicolumn{1}{l|}{28.94}    & \multicolumn{1}{l|}{61.4}     & \multicolumn{1}{l|}{29.97}    & \multicolumn{1}{l|}{20.78}    & \multicolumn{1}{l|}{43.3}     & 20.68    \\ \hline
\end{tabular}
\end{table*}

\begin{figure}[!htbp]
    \centering
    \subfigure[Rate vs. $\alpha$]{
        \begin{minipage}[t]{0.40\columnwidth}
            \centering
            \vspace*{-7pt}
            \includegraphics[width=\columnwidth]{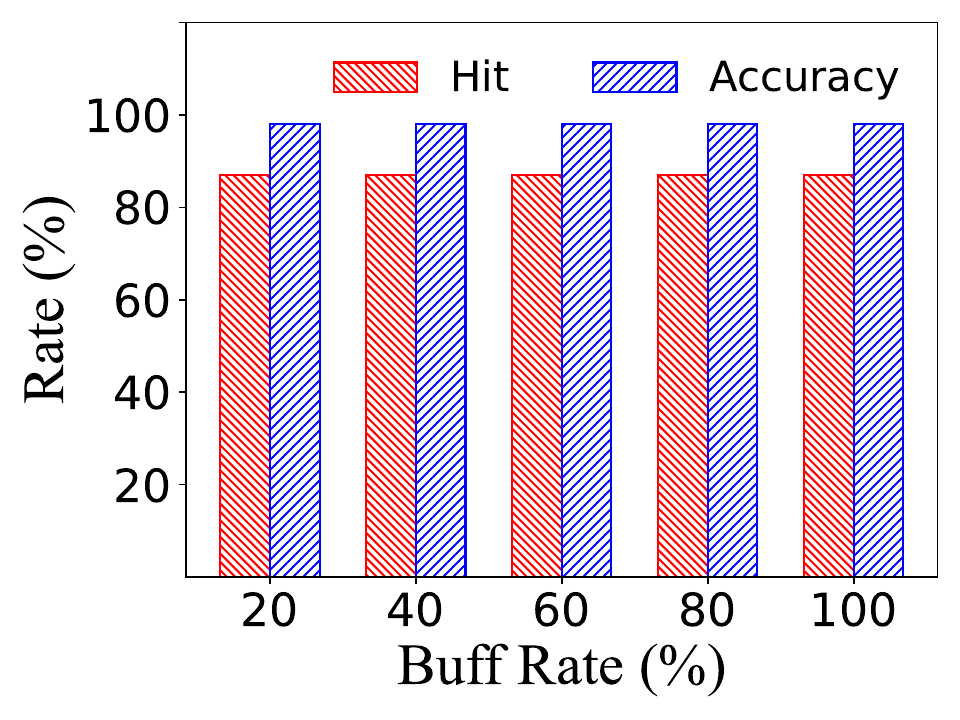}
            \label{fig:bio_Buffrate}
            \vspace*{-10pt}
        \end{minipage}
    }
    \subfigure[Rate vs. $\beta$]{
        \begin{minipage}[t]{0.40\columnwidth}
            \centering
            \vspace*{-7pt}
            \includegraphics[width=\columnwidth]{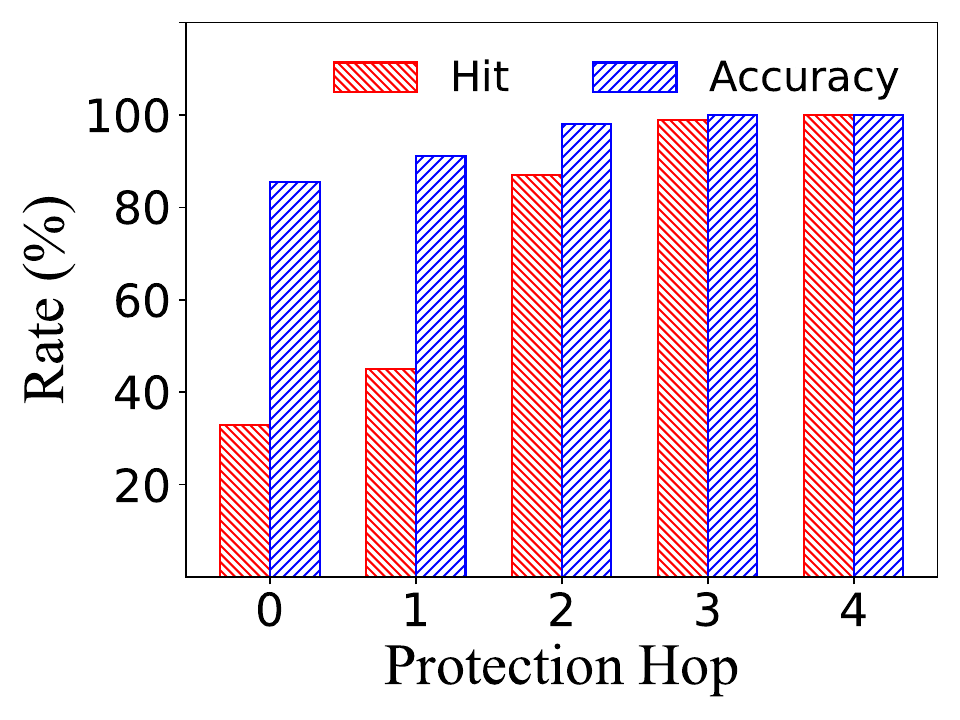}
            \label{fig:bio_InitHop}
            \vspace*{-10pt}
        \end{minipage}
    }
    \centering
    \vspace*{-15pt}
    \caption{Effect of $\alpha$ and $\beta$ (Bio)}
    \label{fig:bio_effect}
\end{figure}

\begin{figure}[!htbp]
    \centering
    \subfigure[Rate vs. $\alpha$]{
        \begin{minipage}[t]{0.40\columnwidth}
            \centering
            \vspace*{-7pt}
            \includegraphics[width=\columnwidth]{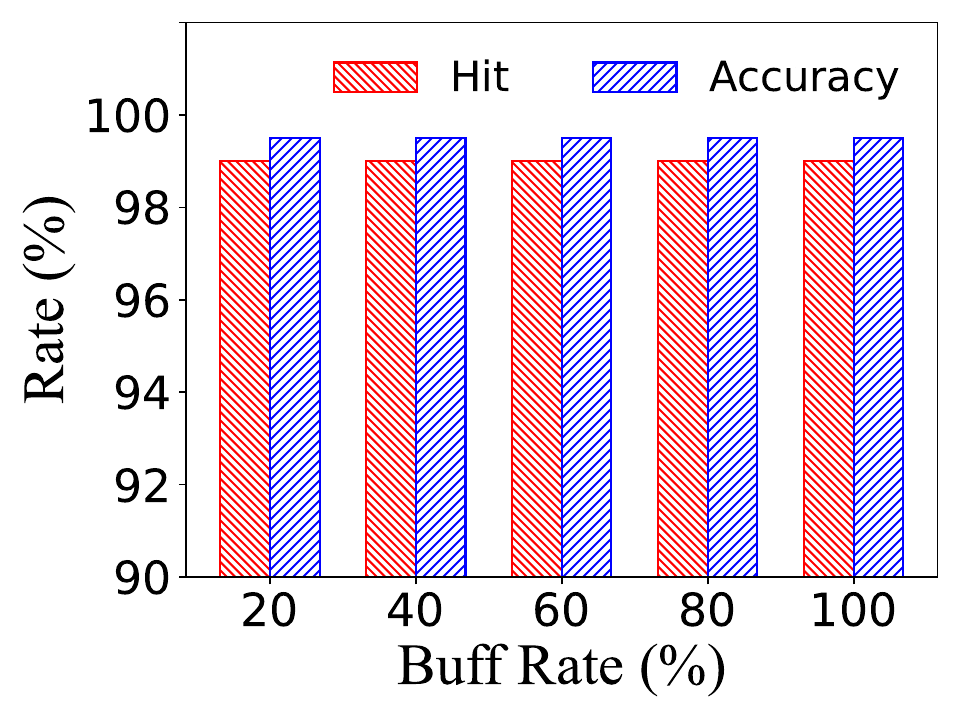}
            \label{fig:web_Buffrate}
            \vspace*{-10pt}
        \end{minipage}
    }
    \subfigure[Rate vs. $\beta$]{
        \begin{minipage}[t]{0.40\columnwidth}
            \centering
            \vspace*{-7pt}
            \includegraphics[width=\columnwidth]{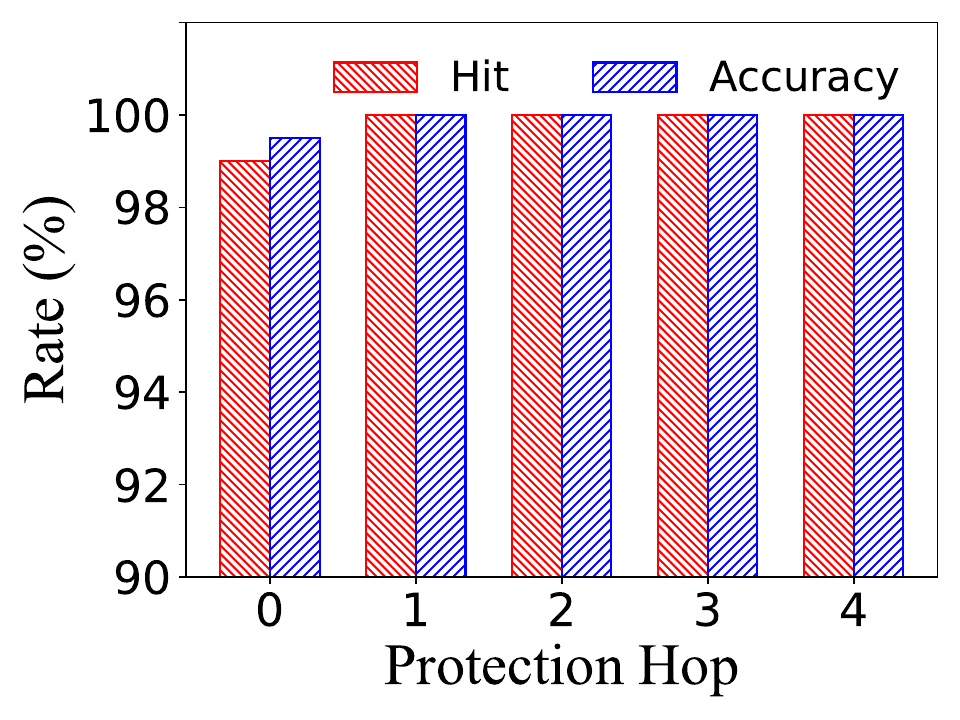}
            \label{fig:web_InitHop}
            \vspace*{-10pt}
        \end{minipage}
    }
    \centering
    \vspace*{-15pt}
    \caption{Effect of $\alpha$ and $\beta$ (Web)}
    \label{fig:web_effect}
\end{figure}

\textbf{Effect of Parameters.} We also investigate the effect of $\textsc{LSearch}$'s parameters, namely buff rate $\alpha$ in Strategy~\ref{strategy:vertex_skip} and protection hop $\beta$ in Strategy~\ref{strategy:protection} (Section~\ref{ssec:LSPS_G_algorithm}).
The parameter settings are listed in Table~\ref{tab:parameter_LSearch} with the default value in bold.
The two parameters have little effect on query time and memory use of $\textsc{LSearch}$, and thus the results are omitted. 
Figure~\ref{fig:bio_effect} shows the effect of $\alpha$ and $\beta$ on hit rate and accuracy on Bio dataset. As the buff rate $\alpha$ increases, both metrics increase slightly. With a larger protection hop $\beta$, the effectiveness of $\textsc{LSearch}$ becomes clearly better, with hit rate improving more rapidly. 
A longer protection hop length $\beta$ means it does not prune any vertex during the expansion within $\beta$, and so it can avoid more unsafe pruning.
Figure~\ref{fig:web_effect} reports the results on Web dataset. As $\alpha$ and $\beta$ increase, both hit rate and accuracy improves gently. When $\alpha$ or $\beta$ is set to a large value, the accuracy reaches $100\%$, meaning the two two strategies enable $\textsc{LSearch}$ to find the groundtruth results. We omit the results on Brain and Power datasets, as they are similar to what we see on Web dataset.

\vspace*{-5pt}
\begin{table}[!htbp]
\small
\caption{Parameter settings of \textsc{LSearch}.}\label{tab:parameter_LSearch}
\vspace*{-10pt}
\begin{tabular}{|l|l|}
\hline
\textbf{Parameter} & \textbf{Setting}  \\ \hline 
$\alpha$ & \textbf{20\%}, 40\%, 60\%, 80\%, 100\% \\ \hline
$\beta$ & \textbf{0}, 1, 2, 3, 4 \\ \hline
\end{tabular}
\end{table}
\vspace*{-10pt}

\subsubsection{Shortest Path Query on Larger Graph}

\begin{figure*}[!htbp]
\centering
    \begin{minipage}[t]{0.4\columnwidth}
        \centering
        \vspace*{-7pt}
        \includegraphics[width=\columnwidth]{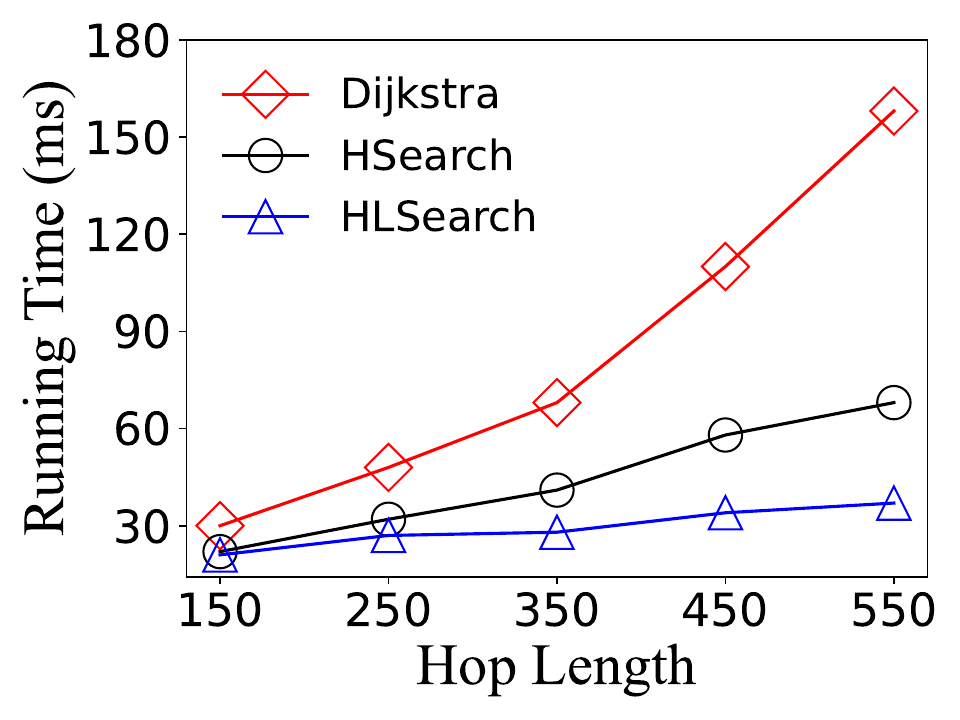}
        \vspace*{-20pt}
        \caption{Time vs. $\rho$}\label{fig:roadNA_HopLength_time}
    \end{minipage}
    \begin{minipage}[t]{0.4\columnwidth}
        \centering
        \vspace*{-7pt}
        \includegraphics[width=\columnwidth]{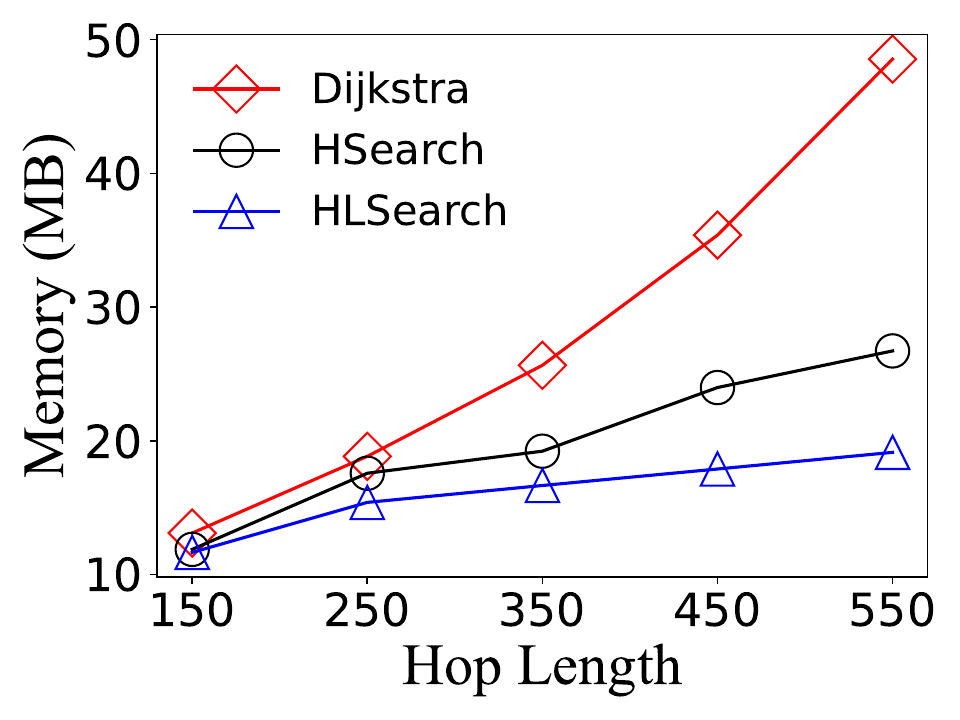}
        \vspace*{-20pt}
        \caption{Mem. vs. $\rho$}\label{fig:roadNA_HopLength_memory}
    \end{minipage}
    \begin{minipage}[t]{0.4\columnwidth}
        \centering
        \vspace*{-7pt}
        \includegraphics[width=\columnwidth]{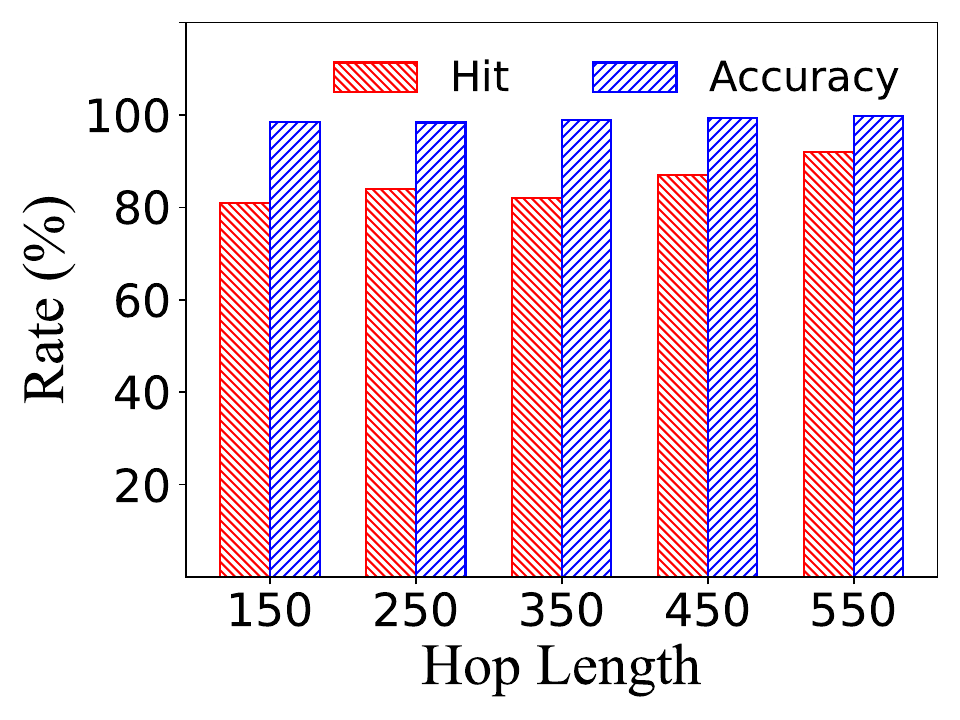}
        \vspace*{-20pt}
        \caption{Rate vs. $\rho$}\label{fig:roadNA_HopLength_effectiveness}
    \end{minipage}
    \begin{minipage}[t]{0.4\columnwidth}
        \centering
        \vspace*{-7pt}
        \includegraphics[width=\columnwidth]{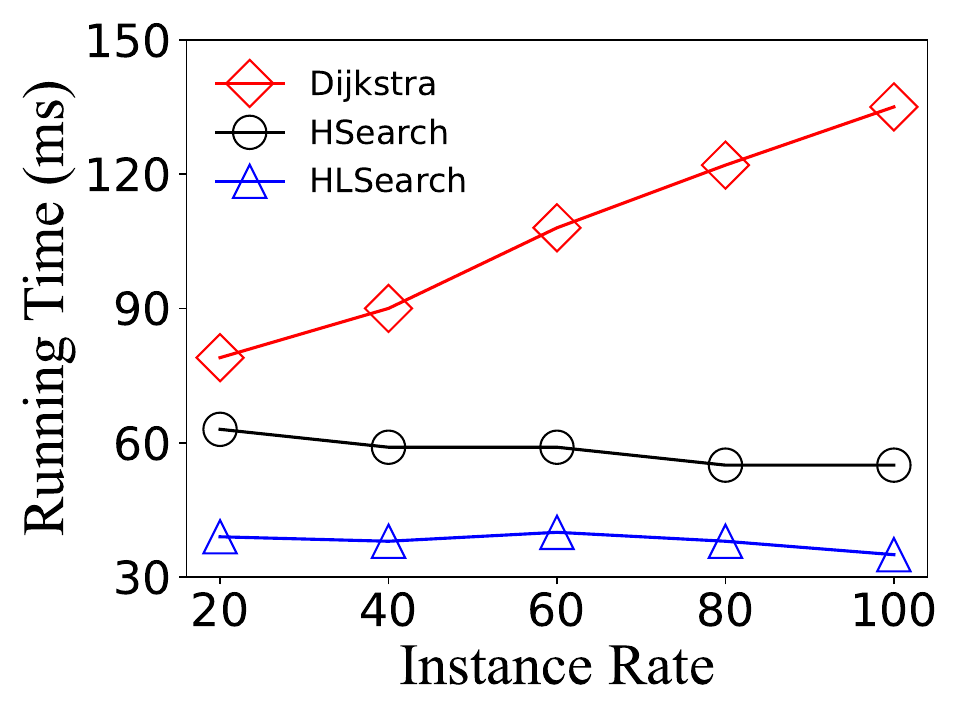}
        \vspace*{-20pt}
        \caption{Time vs. $\eta$}\label{fig:roadNA_InterRate_time}
    \end{minipage}
\end{figure*}

\begin{figure*}[!htbp]
\centering
    \begin{minipage}[t]{0.4\columnwidth}
        \centering
        \vspace*{-7pt}
        \includegraphics[width=\columnwidth]{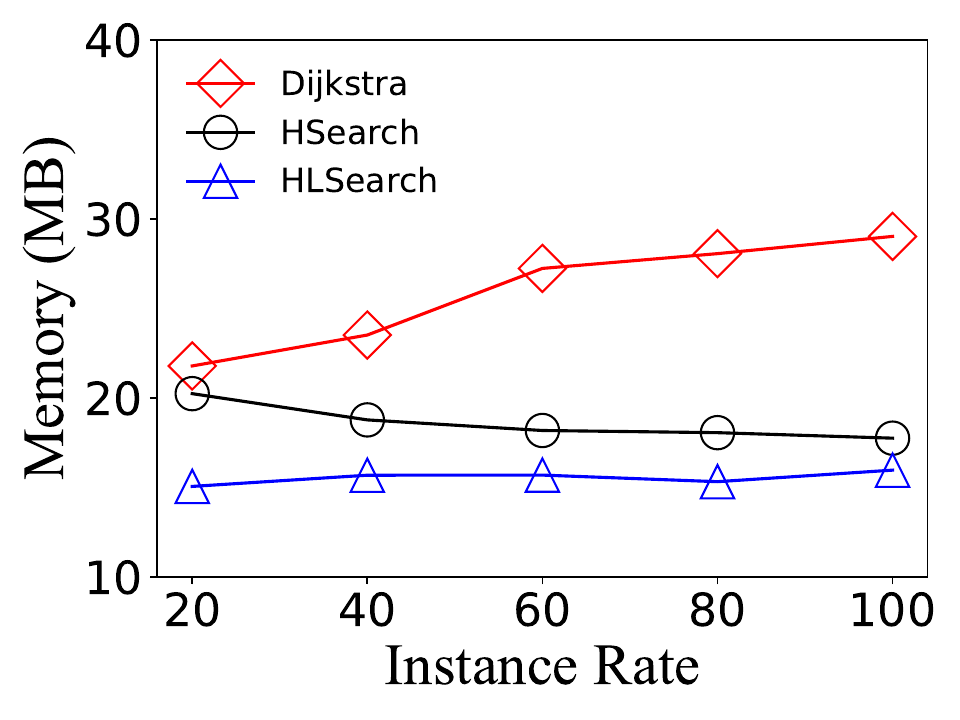}
        \vspace*{-20pt}
        \caption{Mem. vs. $\eta$}\label{fig:roadNA_InterRate_memory}
        \vspace*{-7pt}
    \end{minipage}
    \begin{minipage}[t]{0.4\columnwidth}
        \centering
        \vspace*{-7pt}
        \includegraphics[width=\columnwidth]{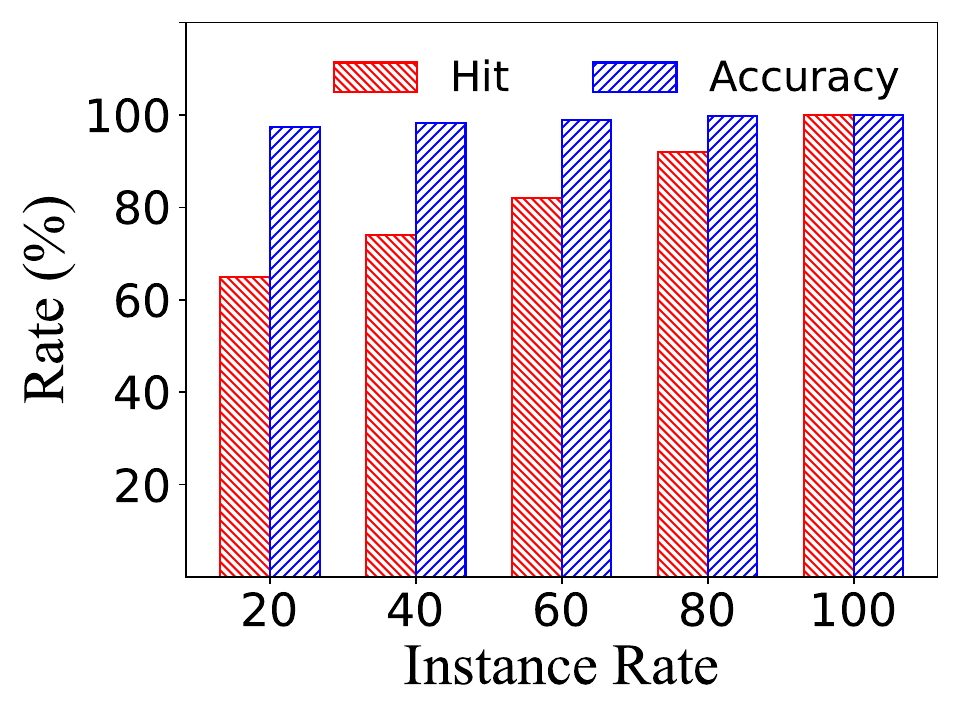}
        \vspace*{-20pt}
        \caption{Rate vs. $\eta$}\label{fig:roadNA_InterRate_effectiveness}
        \vspace*{-7pt}
    \end{minipage}
    \begin{minipage}[t]{0.4\columnwidth}
        \centering
        \vspace*{-7pt}
        \includegraphics[width=\columnwidth]{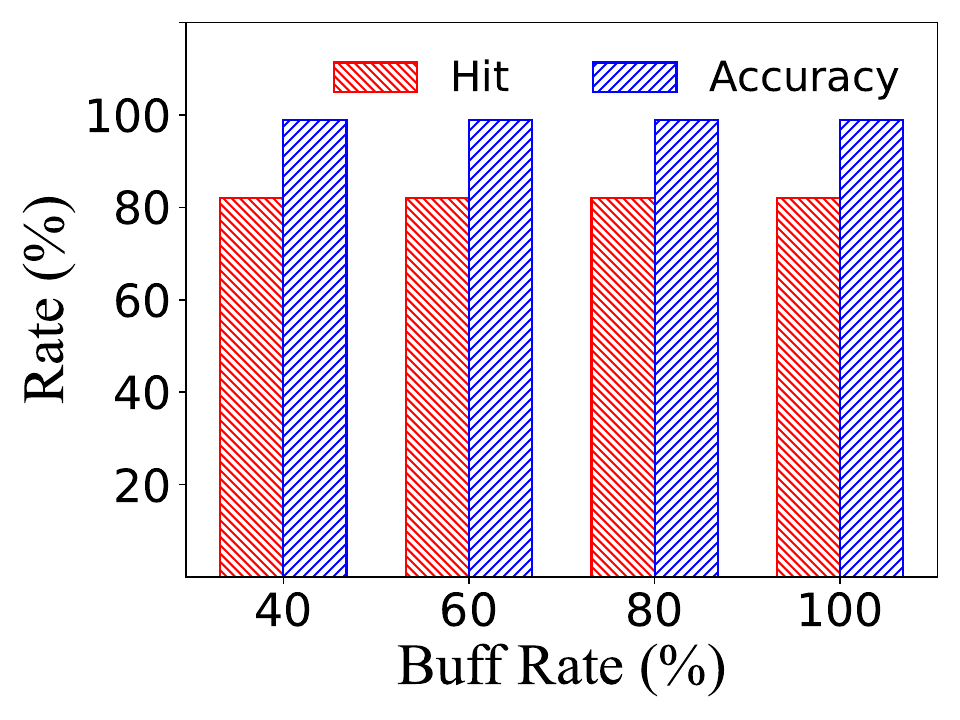}
        \vspace*{-20pt}
        \caption{Rate vs. $\alpha$}\label{fig:roadNA_BuffRate_effectiveness}
        \vspace*{-7pt}
    \end{minipage}
    \begin{minipage}[t]{0.4\columnwidth}
        \centering
        \vspace*{-7pt}
        \includegraphics[width=\columnwidth]{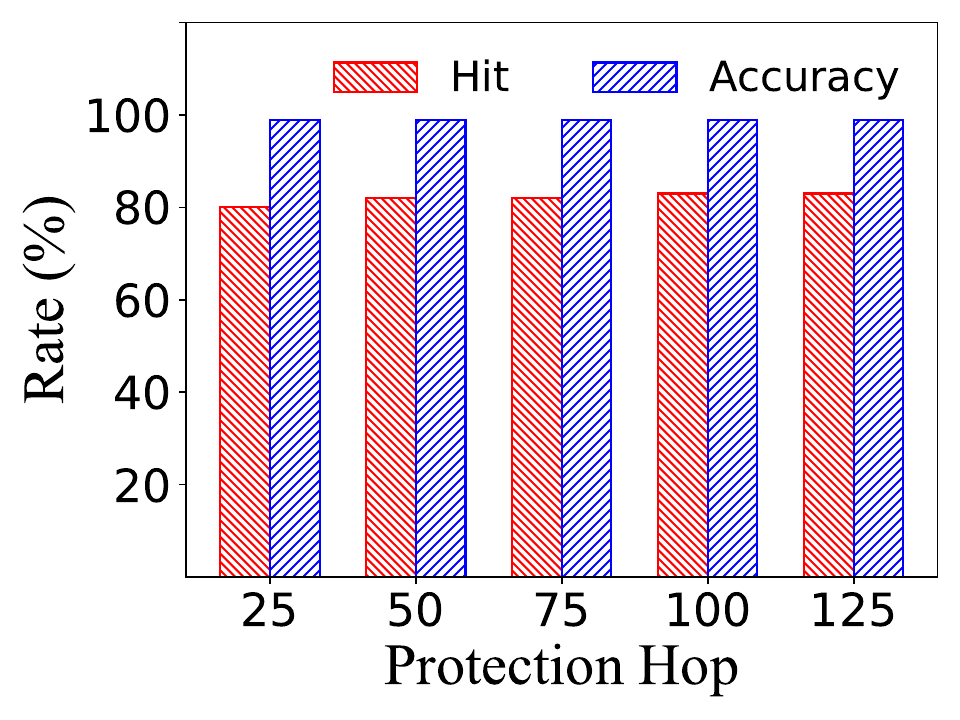}
        \vspace*{-20pt}
        \caption{Rate vs. $\beta$}\label{fig:roadNA_InitHop_effectiveness}
        \vspace*{-7pt}
    \end{minipage}
\end{figure*}

For a larger graph, we construct a hierarchical structure and use $\textsc{HLSearch}$ to query the shortest path. We evaluate $\textsc{HLSearch}$ on the Road-NA dataset.
Since SGNN achieves good performance on Brain, Bio, Web, and Power datasets, where the number of vertex ranges from 503 to 4941, we split Road-NA into a number of subgraphs whose average vertex number is around 4000.
To achieve this, we set the minimum vertex number of each subgraph to 3,500. As a result, the original graph  is split into 43 subgraphs, each forming a leaf node. These leaf nodes are merged into a non-leaf node. Recursively, all such non-leaf nodes are merged into a higher-level non-leaf node until there is only one node. Each non-leaf node contains at least 5 children. After the hierarchical structure is constructed, we prepare the distance matrix for each leaf node and distance-path matrix for each non-leaf node. The overall construction time is around 3 hours.
On top of the hierarchical structure, we also implement a search method $\textsc{HSearch}$ that uses no learning-based strategy. 

The parameter settings of $\textsc{HLSearch}$ are listed in Table~\ref{tab:parameter_HLSearch}. We evaluate the effect of hop $\rho$, instance rate $\eta$, buff rate $\alpha$, and protection hop $\beta$. We random select 100 query instances whose hop of shortest path is between $\rho - 50$ and $\rho + 50$, where $\rho$ is set $150$, $250$, $350$, $450$, and $550$. We find that whether the source vertex and target vertex are from the same leaf node effect the performance of $\textsc{HLSearch}$. Therefore, we use a parameter $\eta$ to control the rate of query instances where the source vertex and target vertex are from different leaf nodes. We also evaluate the effect of $\alpha$ and $\beta$.

\vspace*{-10pt}
\begin{table}[!htbp]
\small
\caption{Parameter settings of \textsc{HLSearch}.}\label{tab:parameter_HLSearch}
\vspace*{-10pt}
\begin{tabular}{|l|l|}
\hline
\textbf{Parameter} & \textbf{Setting}  \\ \hline 
$\rho$ & 150, 250, \textbf{350}, 450, 550 \\ \hline
$\eta$ & 20\%, 40\%, \textbf{60\%}, 80\%, 100\% \\ \hline
$\alpha$ & \textbf{40\%}, 60\%, 80\%, 100\% \\ \hline
$\beta$ & 25, \textbf{50}, 75, 100, 125 \\ \hline
\end{tabular}
\end{table}
\vspace*{-10pt}

\textbf{Effect of $\rho$.} Figures~\ref{fig:roadNA_HopLength_time} and~\ref{fig:roadNA_HopLength_memory} show the effect of $\rho$ for query time and memory use. A larger $\rho$ leads to more time and memory costs for all three methods because more expansions are involved during the search. Among all, $\textsc{HLSearch}$ needs the smallest time and memory cost, as it searches over the hierarchical structure with learning-based strategies. With $\rho$ increasing, the performance gap among the three methods increases gradually. Though $\textsc{HSearch}$ and $\textsc{HLSearch}$ are both relatively stable,  $\textsc{HLSearch}$'s costs are considerably more stable. This is attributed to its learning-based strategies that help prune a large number of unnecessary expansions. 
Since $\textsc{Dijkstra}$ and $\textsc{HSearch}$ are exact methods, we only evaluate the hit rate and accuracy of $\textsc{HLSearch}$. The results are shown in Figure~\ref{fig:roadNA_HopLength_effectiveness}. The distance accuracy is very high, close to 100\%, and the hit rate is larger than 80\%. Again, this shows that $\textsc{HLSearch}$ is quite effective.

\textbf{Effect of $\eta$.} Referring to Figures~\ref{fig:roadNA_InterRate_time} and~\ref{fig:roadNA_InterRate_memory}, when there are more query instances whose source and target are from different leaf nodes, $\textsc{Dijkstra}$ incurs more time and memory costs. When source and target vertices are relatively far apart on the graph, $\textsc{Dijkstra}$ needs to explore more vertices to reach the target vertex. In contrast, with the pre-computed distance information in the hierarchical structure, $\textsc{HSearch}$ and $\textsc{HLSearch}$ can stay stable when $\eta$ is varied. In particular, $\textsc{HLSearch}$ still performs best in terms of query time and memory use. 

Figure~\ref{fig:roadNA_InterRate_effectiveness} shows the results on hit rate and accuracy of $\textsc{HLSearch}$. A larger $\eta$ leads to a higher hit rate and accuracy. When $\eta$ is low, there are more source-target pairs in the same leaf node. In this case, $\textsc{HLSearch}$ searches on a subgraph without using pre-computed information in the hierarchical structure, which however causes more errors in the search results. When $\eta$ increases to 100\%, the hit rate and accuracy are very close to 100\%. In reality, most source and target pairs are far apart and $\eta$ is around 98\%, which means $\textsc{HLSearch}$ can achieve high effectiveness. 

\textbf{Effect of $\alpha$ and $\beta$.} Figures~\ref{fig:roadNA_BuffRate_effectiveness} and~\ref{fig:roadNA_InitHop_effectiveness} show the effects of $\alpha$ and $\beta$ on the effectiveness of $\textsc{HLSearch}$. We can see that the hit rate and accuracy are almost insensitive to $\alpha$ and $\beta$. This is because we pre-compute the distance and connection information between different subgraphs, which in turn helps improve the search effectiveness.

\section{Related Work}
\label{sec:related}


Table~\ref{tab:related_work} summarizes and compares the most relevant shortest path search methods.

\noindent\textbf{Traditional Methods for Shortest Path Problem (SPP).}
Dijkstra algorithm~\cite{dijkstra1959note}, Bellman-Ford algorithm~\cite{bellman1958routing,ford1956network}, A* search algorithm~\cite{hart1968formal}, and Floyd-Warshall algorithm~\cite{floyd1962algorithm} are classical but they are inefficient for large-scale graphs such as modern road networks. 
A number of techniques have been proposed to improve the efficiency of the classical methods, e.g., index-based methods~\cite{sanders2005highway, geisberger2008contraction, hassan2016graph,wang2016effective,ouyang2018hierarchy, ouyang2020efficient, qiu2022efficient}, landmark-based methods~\cite{francis2001idmaps,ng2002predicting,kleinberg2004triangulation,kriegel2008hierarchical,potamias2009fast,gubichev2010fast,qiao2011querying,qiao2012approximate, akiba2013fast}, bidirectional methods~\cite{luby1989bidirectional,geisberger2008contraction,vaira2011parallel,cabrera2020exact}, and graph compression~\cite{qiao2011querying}. 
The existing surveys~\cite{bienstock1991graph, madkour2017survey, yan2020comprehensive} summarize relevant techniques.
All these traditional speed-up techniques either incur huge pre-processing time or need large space to store the indexes built before any search. Moreover, most of them are designed for a single type of graph, namely road networks with special features like longitude and latitude.

\noindent\textbf{Learned Methods for Path Search.}
In recent years, there has been a growing interest in using machine learning techniques to improve the efficiency and accuracy of shortest path search algorithms. Li et al.~\cite{li2024reinforcement} propose the ReinforceRouting model to generate evacuation routes on street networks. Qi et al.~\cite{qi2020learning} learn an embedding for the graph and train a multi-layer perceptron (MLP) to predict the distance between node pairs given their embeddings. Aiming at SSP on a large dynamic graph, Yin et al.~\cite{yin2021learning} propose a learning framework, which decomposes a large SPP instance into multiple small instances and learns GCN-DQN models to solve small shortest path problem instances. Huang et al.~\cite{huang2021learning} learn a road network embedding (RNE) model and use a hierarchical learning strategy to compute an approximate shortest-path distance. Zhao et al.~\cite{zhao2022rne} further entend the RNE model to compute an approximate shortest path over a road network.
Some of these methods return only shortest path distances but not the paths; and all of them need extra graph features like longitude, latitude, and road categories to enable the learning. Consequently, these methods are inapplicable to our problem setting of generic graphs.
In this work, our proposed SGNN can predict the shortest distance without using any context information and thus it can work on all types of graphs. 

\vspace*{-5pt}
\begin{table}[!htbp]
\small
\caption{Overview of related works.} \label{tab:related_work}
\vspace*{-10pt}
\begin{tabular}{|c|c|c|c|c|c|}
\hline
\textbf{Type}                   & \textbf{Name}  & \textbf{Year} & \textbf{Path} & \textbf{Graph} & \textbf{Result} \\ \hline
\multirow{2}{4em}{Classical}        & Dijkstra~\cite{dijkstra1959note}       & 1959          & Yes           & Genetic                 & Exact                      \\ \cline{2-6} 
                                & A*~\cite{hart1968formal}             & 1968          & Yes           & Genetic                 & Approx.                \\ \hline
\multirow{6}{4em}{Index- based}    & HH~\cite{sanders2005highway}             & 2005          & Yes           & Spatial       & Exact                      \\ \cline{2-6} 
                                & CH~\cite{geisberger2008contraction}             & 2008          & Yes           & Spatial       & Exact                      \\ \cline{2-6} 
                                & COLA~\cite{wang2016effective}           & 2016          & Yes           & Spatial       & Approx.                \\ \cline{2-6} 
                                & H2H~\cite{ouyang2018hierarchy}            & 2018          & Yes           & Spatial       & Exact                      \\ \cline{2-6} 
                                & Shortcut Index~\cite{ouyang2020efficient} & 2020          & Yes           & Spatial       & Approx.                \\ \cline{2-6} 
                                & TL-Index~\cite{qiu2022efficient}       & 2022          & Yes           & Spatial       & Exact                      \\ \hline
\multirow{3}{4em}{Landmark- based} & A*+RNE~\cite{kriegel2008hierarchical}         & 2008          & Yes           & Spatial       & Exact                      \\ \cline{2-6} 
                                & LLS~\cite{qiao2012approximate}            & 2012          & Yes           & Genetic                 & Approx.                \\ \cline{2-6} 
                                & PLL~\cite{akiba2013fast}            & 2013          & No            & Genetic                 & Exact                      \\ \hline
\multirow{5}{4em}{Learning- based} & vdist2vec~\cite{qi2020learning}      & 2020          & No            & Spatial       & Approx.                \\ \cline{2-6} 
                                & SPP-GS~\cite{yin2021learning}         & 2021          & Yes           & Spatial       & Approx.                \\ \cline{2-6} 
                                & RNE~\cite{huang2021learning}            & 2021          & No            & Spatial       & Approx.                \\ \cline{2-6} 
                                & RNE+~\cite{zhao2022rne}           & 2021          & Yes           & Spatial       & Approx.                \\ \cline{2-6}
                                & ReinforceRouting~\cite{li2024reinforcement}           & 2024          & Yes           & Spatial       & Approx.                \\
                                
                                \hline
\end{tabular}
\end{table}
\vspace*{-10pt}

\section{Conclusion and Future Work}
\label{sec:conclusion}

In this paper, we study learning-based methods for shortest path search on generic graphs. 
Our contributions include proposing a Skeleton Graph Neural Network (SGNN) for predicting distances and hop lengths on such graphs, and developing a learning-based shortest path search algorithm (\textsc{LSearch}) with learning-based pruning strategies. To handle larger graphs, we design a hierarchical structure and develop \textsc{HLSearch} for finding shortest paths on it. We evaluate our methods through extensive experiments on real-world graph datasets, and demonstrate that they outperform state-of-the-art methods in terms of efficiency and effectiveness.

For future work, it is possible to explore adapting SGNN to dynamic graphs that require model updates. Also, it is worthwhile to develop rigorous error bounds for learning-based path search. Moreover, it is relevant to use the methods in this paper as building blocks to make advanced path planning problems more efficient.

\begin{acks}
 This work was supported by Independent Research FundDenmark (No. 1032-00481B).
\end{acks}

\clearpage

\newpage
\balance
\bibliographystyle{ACM-Reference-Format}
\bibliography{ref}

\end{document}